\documentclass{article}
\usepackage{enumitem}
\usepackage{arxiv}

\usepackage[utf8]{inputenc} % allow utf-8 input
\usepackage[T1]{fontenc}    % use 8-bit T1 fonts
\usepackage{booktabs}       % professional-quality tables
\usepackage{nicefrac}       % compact symbols for 1/2, etc.
\usepackage{microtype}      % microtypography
\usepackage{lipsum}
\usepackage{multirow}
\usepackage[table]{xcolor}
\usepackage{arydshln}
\usepackage{siunitx}
\usepackage{makecell} % For line breaks in cells
\usepackage{mathrsfs}
\usepackage{array}
\usepackage{svg}
\usepackage{algpseudocode}
\usepackage{algorithm}
\usepackage{array}
\usepackage[caption=false,font=normalsize,labelfont=sf,textfont=sf]{subfig}
\usepackage{textcomp}
\usepackage{stfloats}
\usepackage{url}
\usepackage{verbatim}
\usepackage{rotating}
\usepackage{graphicx}
\usepackage{multicol}
\usepackage{cellspace}
\usepackage{subcaption}
\usepackage{amsmath, amsfonts, amsthm, amssymb}
\usepackage{fancyhdr}
\usepackage{hyperref}
\usepackage{enumitem}
\usepackage{wrapfig}
\usepackage{pdflscape}   % for landscape orientation
\usepackage{longtable}   % for multi-page tables
\usepackage{multirow}    % if you need multirow (not used 
\usepackage{afterpage}   % to insert a landscape page cleanly

\title{Paper08-IEEE}
\author{itxwaleedrazzaq }
\date{}

\theoremstyle{definition}

\newcounter{problem}

\theoremstyle{remark}

\title{\Large Continuous-Time Machine Learning: A Unified Mathematical Perspective}

\author{
  Waleed Razzaq\thanks{Corresponding authors: \texttt{waleedrazzaq@mail.ustc.edu.cn} ;  \texttt{ybzhao@ustc.edu.cn}} \\
  School of Automation\\
  University of Science and Technology of China\\
  Hefei, Anhui \\
  \texttt{waleedrazzaq@mail.ustc.edu.cn} \\
  \And
  Yun-Sheng Zhao \\
  School of Automation\\
  University of Science and Technology of China\\
  Hefei, Anhui \\
  \texttt{zys1030@mail.ustc.edu.cn} \\
  \And
  Yun-Bo Zhao$^{*}$ \\
  School of Automation\\
  University of Science and Technology of China\\
  Hefei, Anhui \\
  \texttt{ybzhao@ustc.edu.cn} \\
}

\begin{document}
% \scriptsize
\maketitle

\begin{abstract}
Continuous-time (CT) machine learning has emerged as a principled framework for modeling temporal dynamics as a continuous process, particularly when observations are sampled at arbitrary time  points or span long-range horizons. However, major branches of CT machine learning have matured in separate research communities, leaving their mathematical relationships and design trade-offs insufficiently characterized. In this survey, we develop a unified, concept-driven view of major CT machine learning branches through a taxonomy that organizes families according to their underlying base mathematical formulations. We present a canonical mathematical formulation that relates these families through different architectural choices of vector-field parameterization, stochasticity, memory mechanisms, and discretization. We compare training algorithms, optimization strategies, and failure modes, highlighting the trade-offs across families. We further provide a comparative analysis of theoretical computational complexity alongside an illustrative architecture-controlled benchmark analysis on representative architectures from each family. We also review software ecosystems supporting their implementation. Finally, we identify open challenges in approximation theory, training stability, hardware-efficient implementations, benchmarking, foundation models, and scientific machine learning, and discuss an agenda for future research. 
\end{abstract}
% keywords can be removed

\section{Introduction}
Real-world data are often observed at irregular spaced points in time, while underlying processes evolve continuously and may exhibit dependencies over multiple timescales. Conventional discrete-time (DT) machine learning methods typically rely on a predefined time grid, requiring resampling or interpolation for irregular observations and potentially obscuring the underlying dynamics. These limitations have motivated continuous-time (CT) machine learning that models temporal dynamics as continuous processes. Different research communities have developed distinct approaches to represent and learn these temporal evolutions. Neural differential equations (NDEs)~\cite{chen2018neural, rubanova2019latent, lechner2022mixed, kidger2020neural, tzen2019neural, li2020scalable, kong2020sde} emerged primarily from the deep learning community. State-space models (SSMs) draw on ideas from control theory, numerical analysis, and systems identification~\cite{gu2020hippo, gu2021efficiently, gu2022parameterization}. Biologically-Inspired Liquid Networks (BI-LNs)~\cite{lechner2018neuronal, hasani2021liquid, hasani2018liquid, hasani2022closed,liang2026rederived} are rooted in computational neuroscience. CT-Attention~\cite{chien2021continuous, zhang2025continuous, razzaq2025neuronal} or CT-Transformers~\cite{chen2023contiformer, d2023odeformer, kan2025ot, razzaq2026fluid} are largely motivated by the challenges of learning representative patterns from long-range, irregularly sampled temporal data. Despite pursuing a common objective, these communities have largely evolved in isolation, adopting distinct theoretical perspectives, evaluation protocols, and benchmark datasets, resulting in limited integration of insights across communities. \\
Prior surveys typically review an individual branch or a subset of branches rather than providing a unified treatment across communities. Surveys on NDEs~\cite{kidger2022neural,oh2025comprehensive} review only models where time ($t$) serves as an integration bound rather than an explicit independent variable and exclude other branches. Surveys on SSMs~\cite{somvanshi2025s4, tiezzi2025state} fail to engage with BI-LNs and CT-Transformers. Reviews on BI-LNs~\cite{zong2025accuracymemoryefficiencygeneralization, jammal2025comparative,Wahidi2025ScaleIN} focus on their scalability and surrogate gradients but miss the structural connections to SSMs and NDEs. Surveys on transformers~\cite{wen2022transformers, sommers2024survey} focus primarily on DT sequence modeling and largely predate recent advances in CT Transformer architectures. As a result, these architectures have received limited survey coverage. Computational trade-offs among branches are rarely compared, and theoretical synthesis is largely absent. Existing surveys rarely make explicit the mathematical connections across branches or critically analyze why one branch succeeds where another fails.
\begin{wrapfigure}{r}{0.6\textwidth}
\vspace{-1mm}
\captionsetup{type=table}
\centering
\caption{Comparison of existing surveys}
\vspace{1mm}
\label{tab:survey_comparison}
\resizebox{0.57\textwidth}{!}{%
\begin{tabular}{cccccc}
\toprule
\textbf{Survey Type} & Reference & NDEs & SSMs & BI-LNs & CT-Transformers \\
\midrule

NDEs & \cite{kidger2022neural,oh2025comprehensive} & \checkmark & $\times$ & $\times$ & $\times$  \\

SSMs & \cite{somvanshi2025s4,tiezzi2025state} & $\times$ & \checkmark & $\times$ & $\times$ \\

BI-LNs & \cite{jammal2025comparative, zong2025accuracymemoryefficiencygeneralization,Wahidi2025ScaleIN}, & $\times$ & $\times$ & \checkmark & $\times$  \\

Transformers & \cite{wen2022transformers,sommers2024survey} & $\times$ & $\times$ & $\times$ & $\times$ \\
\midrule
\textbf{Ours} & -- & \checkmark & \checkmark & \checkmark & \checkmark\\
\bottomrule
\end{tabular}}
\begin{minipage}{0.57\textwidth}
\footnotesize
\textbf{Note:} \checkmark indicates that the branch's core models are technically analyzed (e.g., formulations, training, or benchmarks); \(\times\) indicates that the branch is absent or mentioned only in passing.

\end{minipage}
\vspace{-7pt}
\end{wrapfigure}
To address these limitations, we present a concept-driven taxonomy that organizes CT machine learning into five families: (i) \textit{Linear Dynamical Systems}; (ii) \textit{Linearly-Coupled Dynamics with State-Dependent Decay}; (iii) \textit{Freely Parameterized Vector Fields}; (iv) \textit{Selective Scan State-Space Models}; and (v) \textit{Continuous-Time Transformers}, by the constraints they impose on a common canonical formulation. We also make explicit how these families can be interpreted through different architectural choices of vector-field parameterization, stochasticity, memory mechanisms, and discretization on the canonical formulation. We also discuss training algorithms, optimization strategies, and failure modes among families. We then provide a comparative study on theoretical computational complexity with illustrative architecture-controlled benchmark analysis and software ecosystem. Finally, we highlight open challenges in approximation theory, training stability, hardware-efficient implementation and propose a research agenda that connects CT learning to foundation models, scientific computing, and autonomous systems. Table~\ref{tab:survey_comparison} compares the existing surveys against this survey.

% \begin{table}[ht!]
% \centering
% \caption{Comparison of existing surveys}
% \vspace{1mm}
% \label{tab:survey_comparison}
% \resizebox{0.65\textwidth}{!}{%
% \begin{tabular}{cccccc}
% \toprule
% \textbf{Survey Type} & Reference & NDEs & SSMs & BI-LNs & CT-Transformers \\
% \midrule

% NDEs & \cite{kidger2022neural,oh2025comprehensive} & \checkmark & $\times$ & $\times$ & $\times$  \\

% SSMs & \cite{somvanshi2025s4,tiezzi2025state} & $\times$ & \checkmark & $\times$ & $\times$ \\

% BI-LNs & \cite{jammal2025comparative, zong2025accuracymemoryefficiencygeneralization,Wahidi2025ScaleIN}, & $\times$ & $\times$ & \checkmark & $\times$  \\

% Transformers & \cite{wen2022transformers,sommers2024survey} & $\times$ & $\times$ & $\times$ & $\times$ \\
% \midrule
% \textbf{Ours} & $\times$ & \checkmark & \checkmark & \checkmark & \checkmark\\
% \bottomrule
% \end{tabular}}
% \end{table}

\begin{wrapfigure}{r}{0.53\textwidth}
\vspace{-10mm}
\captionsetup{type=table}
\centering
\caption{Notations used throughout the survey.}
\vspace{1mm}
\label{tab:notation}
\resizebox{0.52\textwidth}{!}{%
\begin{tabular}{lll}
\toprule
\textbf{Variable} & \textbf{Continuous} & \textbf{Discrete} \\
\midrule
Time & $t$ & $t_k$ \\
Initial / terminal time & $t_0,\ t_L$ & $\times$ \\
Hidden state & $x(t) \in \mathbb{R}^{N}$ & $x_k$ \\
Input & $u(t) \in \mathbb{R}^{d_{in}}$ & $u_k$ \\
Output & $y(t) \in \mathbb{R}^{d_{out}}$ & $y_k$ \\
Wiener process & $\mathcal{W}(t) \in \mathbb{R}^{m}$ & $\times$ \\
Time step & $\times$ &  $\Delta_k$ \\
Time constant & $\tau(x,u)$ & $\tau(x,u)$ \\
State transition matrix & $A$ & $\bar{A}_k = e^{A(\Delta_k)}$ \\
Input matrix & $B$ & $\bar{B}_k =
\int_{0}^{\Delta_k}
e^{A(\Delta_k-\tau)}B\,d\tau$ \\
Output matrix & $C$ & $C_k$ \\
Feedthrough matrix & $D$ & $D_k$ \\
Drift vector field & $f(x(t),u(t),t;\theta)$ & $f(x_k,u_k;\theta)$ \\
Diffusion field & $g(x(t),u(t),t;\varphi)$ & $g(x_k,u_k;\varphi)$ \\
Readout map & $h(x(t),u(t);\psi)$ & $h(x_k,u_k;\psi)$ \\
Model parameters & $\theta,\ \varphi,\ \psi$ & $\theta,\ \varphi,\ \psi$ \\
Discretization operator & $\times$ & $\mathcal{D}(x_{k-1},u_k,\Delta_k;\theta)$ \\
\midrule
\multicolumn{3}{l}{\textit{Operators and conventions}}\\
\midrule
Time derivative & $\dot{x} = \tfrac{dx}{dt}$ & $\times$ \\
Elementwise (Hadamard) product & $\odot$ & $\odot$ \\
Activation function & $\sigma(\cdot)$ & $\sigma(\cdot)$ \\
Standard operators & \multicolumn{2}{l}{$\mathrm{diag}(\cdot),\,\ \mathrm{softmax}(\cdot),\ \mathrm{softplus}(\cdot)$} \\
\bottomrule
\end{tabular}}
\begin{minipage}{0.52\textwidth}
\footnotesize
\textbf{Note:} \(\times\) indicates not available.
\end{minipage}
\vspace{-12mm}
\end{wrapfigure}

\subsection{Notations and Mathematical Conventions}
Throughout the survey, we adopt a consistent mathematical notation to emphasize common structures shared across families. Table~\ref{tab:notation} summarizes the notation in its continuous and discrete counterparts.

\subsection{Scope Boundaries}

We explicitly define the scope of this survey to clarify its focus and explain why certain mathematically adjacent branches are not considered in detail.

\textbf{\textit{In Scope:}} Models are considered in scope if they are designed for temporal sequence modeling and their computational formulation explicitly incorporates CT dynamics, with time \emph{(t)} serving as a computational variable, or if their discrete counterparts contain update mechanisms that can be interpreted as discretized approximations. This includes NDEs, SSMs, BI-LNs, and CT-Attention or CT-Transformer variants in which query/key/value construction, positional representation or attention-logit computation are explicitly parameterized by \emph{t}. DT architectures are only included when their state-update rule exhibits a mathematically meaningful correspondence to CT evolution, even when the original model is not explicitly formulated as a CT system.

\textbf{\textit{Out of Scope:}} Models are excluded when their computation has no explicit CT formulation, no meaningful discretization or approximation of CT dynamics, and no identifiable computational mechanism that corresponds to CT evolution. Generative models that have an explicit CT formulation, such as diffusion, score-based, and stochastic-interpolant models, are also excluded and may be cited only for contrast or contextual reference. A fully comprehensive benchmark covering all in-scope architectures, variants, and implementation configurations is also beyond the scope of this survey, as our primary focus is on analyzing and organizing based on their mathematical dynamics.

\subsection{Methodology}

In this subsection, we will briefly describe the methodology adopted throughout the survey.

\textit{\textbf{Sources:}} We searched arXiv across all categories, focusing in particular on cs.LG, cs.AI, cs.CL, cs.SY, and cs.NE. We also examined the proceedings of NeurIPS, ICML, ICLR, AAAI, KDD, AISTATS, UAI, EMLNP, IJCAI, ACL, and workshops, together with relevant IEEE/Elsevier/Springer/ACM venues, namely TNNLS, TAI, TKDE, TAC, TPAMI, Neural Networks, etc. We further expanded the search on Google Scholar by following backward and forward citations from the anchor papers used to develop our taxonomy.

\textbf{\textit{Strategy:}} Our search used two complementary sources of evidence: (a) list of terms covering the main branches of CT temporal modeling: {“continuous-time”, “neural ordinary differential equation”, neural controlled differential equation”, “neural rough differential equation”, “state-space model”, “liquid neural networks”, “liquid time-constant”, “continuous-time attention”, “continuous-time transformers”}; (b) the qualifier list {“survey”, “review”, “overview”} to identify existing surveys and reviews.

\textbf{\textit{Eligibility: }}We included (a) peer-reviewed papers and publicly available preprints up to August 2026, with a verifiable identifier, such as a DOI or arXiv ID; (b) papers that propose, analyze, or benchmark a novel temporal model with an explicit CT formulation 
(compliance with scope); (c) previous surveys and reviews from the past 3-4 years, which we considered separately for the comparison in Table~\ref{tab:survey_comparison}. Preprints are cited using their arXiv identifiers and version information. When a preprint has appeared in a peer-reviewed venue, we cite the published version. We excluded (a) workshop notes that do not provide enough details to verify the model; (b) application-specific tailored hybrid models; (c) preprints that contain no substantive new material.

\subsection{Roadmap}

The remainder of this survey is organized as follows. Section~\ref{section:history} reviews the historical development. Section~\ref{section:canonical} presents a concept-driven taxonomy by organizing families by their base mathematical equations and demonstrating how these families can be interpreted from the canonical dynamical system through different architectural choices. Section~\ref{section:training} discusses training algorithms, optimizations, and trade-offs among families. Section~\ref{section:benchmarks} provides a comparative analysis of computational complexity, alongside a illustrative architecture-controlled benchmark analysis and the supporting software ecosystem. Section~\ref{section:open_problem} identifies open challenges across mathematical, algorithmic, computational, and theoretical aspects. Section~\ref{section:agenda} outlines a future research agenda. Section~\ref{section:conclusion} concludes the survey.
\section{Historical Evolution}\label{section:history}

In this section, we will briefly review the historical development of CT machine learning.

\subsection{Continuous-Time Neural Dynamics}
The concept of modeling neural dynamics as a continuous process is nearly as old as machine learning research itself and has remained a recurring theme throughout its evolution. McCulloch and Pitts~\cite{mcculloch1943logical} introduced binary threshold neurons where neural activity evolves through synchronous discrete updates, and the timing of neural firing is not a computational variable. Hodgkin and Huxley~\cite{hodgkin1952quantitative} developed the first widely recognized CT mathematical model of neural dynamics, describing action potential generation through a system of four coupled nonlinear differential equations governing membrane potential and ion-gate dynamics. Their model was a biophysical description rather than a machine learning architecture, but it established the fundamental principle that neural computation unfolds in continuous time. Hopfield et al.~\cite{Hopfield1984, hopfield1985neural} introduced a CT formulation of neural networks through a circuit-based implementation using amplifiers, resistors, and capacitors. The resulting dynamics converged to fixed points in an energy landscape~\cite{cohen1983absolute}, demonstrating that a CT system could solve optimization problems, but the framework was limited to convergent attractor dynamics.\\
Funahashi and Nakamura~\cite{Funahashi1993} established that CT-RNNs are universal approximators of smooth flows, extending Cybenko’s universal approximation theorem~\cite{cybenko1989approximation} to dynamical systems. This showed that the finite-time trajectories of any smooth dynamical system can be approximated arbitrarily well by a sufficiently large CT-RNN. Consequently, a CT-RNN governed by a first-order ODE with a leak term, a recurrent nonlinearity, and an input drive emerged as the canonical form for CT neural computation.  Beer~\cite{beer1995dynamics} analyzed the bifurcation structure of small CT-RNNs, demonstrating that networks with as few as two neurons can exhibit qualitatively distinct dynamical behaviors, including: (a) fixed points; (b) limit cycles; and (c) bistability. He further demonstrated that the boundaries separating these dynamical regimes vary continuously with the model parameters, indicating that gradual parameter changes can produce predictable transitions between qualitative behaviors. This analysis established an important theoretical foundation for understanding how training shifts networks across dynamical regimes. The emergence of the Elman RNN~\cite{elman1990finding}, trained using backpropagation through time (BPTT)~\cite{rumelhart1986learning}, established a parallel DT model. It can be understood as an Euler discretization of a CT-RNN with unit time constant and step-size~\cite{mozer2017discrete}, but this mathematical relationship remained largely unrecognized for decades, and the two communities developed independently.

\subsection{Gated Recurrent Networks}
Hochreiter and Schmidhuber~\cite{hochreiter1997long} introduced the long-short-term-memory (LSTM) network to address optimization difficulties in RNNs, not as an explicitly CT architecture. The LSTM’s cell state update with its forget and input gates can be reinterpreted as an adaptive time-constant integration. The forget gate controls how much of the previous state is retained, analogous to the leak rate in CT-RNNs. When the forget gate is near 1, the time constant is effectively infinite, and the cell integrates input over long periods. When it is near 0, the state resets rapidly. This CT perspective was recognized only later when subsequent study~\cite{mozer2017discrete} showed that gating mechanisms are implicit ways to learn time constants from data. The gated-recurrent-unit (GRU)~\cite{cho2014learning} simplified the gating structure of the LSTM with only two gates: (a) reset; and (b) update, while essentially preserving the adaptive-time-constant property. However, neither architecture treated time as a continuous variable; time was evaluated only at discrete observation times.

\subsection{Neural Differential Equations (NDEs)}
Chen et al.~\cite{chen2018neural} introduced neural ordinary differential equation (NODE), representing a major conceptual paradigm shift in CT machine learning by parameterizing the vector field of the latent dynamics by an arbitrary neural network, freeing the model from the rigid functional form of CT-RNNs. The adjoint sensitivity method~\cite{chen2018neural, zhuang2020adaptive} adapted from optimal control theory enabled memory-efficient training by computing gradients without storing intermediate solver states. NODE was positioned as a continuous-depth generalization of residual networks~\cite{he2016deep}. The Latent-ODE~\cite{rubanova2019latent} extended the framework for irregularly sampled time series by encoding observation sequences into an initial latent state, evolving that state with an ODE, and decoding it to reconstruct the trajectory. Lechner and Hasani~\cite{lechner2022mixed} proposed ODE-LSTM, a hybrid architecture where the latent state evolves continuously via an ODE but is updated discontinuously when a new observation arrives, further bridging the gap between discrete and continuous. Two important limitations of NODE were recognized immediately. \textit{First}, the input $u(t)$ must be defined as a continuous function of \emph{t}, requiring an external interpolation step for discrete observations. Neural controlled differential equations (NCDE)~\cite{kidger2020neural} addressed this by incorporating the input path directly into the dynamics through a Riemann-Stieltjes integral, eliminating the need for separate interpolation. \textit{Second}, purely deterministic dynamics cannot capture intrinsic stochasticity in the data. Neural stochastic differential equation (NSDEs)~\cite{tzen2019neural, li2020scalable} introduced a diffusion term, jointly quantifying both aleatoric and epistemic regularization through a Bayesian treatment~\cite{kong2020sde}.

\subsection{State-Space Models (SSMs)}
State-space models (SSMs) are fundamental tools in control theory and signal processing, but their integration with modern machine learning was pioneered by the Legendre Memory Unit (LMU)~\cite{voelker2019legendre}. LMU introduced a structured state-space memory mechanism based on linear dynamics with fixed transition matrices derived from Legendre polynomial representations~\cite{szeg1939orthogonal}. By encoding recent input history into a compact latent state, LMU provided a principled alternative to unconstrained recurrent memory mechanisms. Although initially developed within a Nengo neural engineering framework~\cite{Bekolay2014}, the LMU was not recognized as a mainstream machine learning architecture. \\
The HiPPO~\cite{gu2020hippo} framework provides a general formulation for online compression of an input history by projecting it onto a polynomial basis under a specified measure. The choice of measure determines which parts of the past are emphasized. In particular, HiPPO-LegS uses a scaled Legendre measure that represents the full history, while other HiPPO variants induce different temporal weighting. Thus, the long-range memory behavior of HiPPO is a consequence of its structured history representation and projection dynamics rather than the eigenvalues of the transition matrix alone. S4~\cite{gu2021efficiently} combines HiPPO-derived structure with a diagonal-plus-low-rank (DPLR) parameterization, while the resulting SSM parameters are learned during training. Mamba~\cite{gu2023mamba} makes the SSM parameters input-dependent, including the discretization step and input/output projections, enabling selective propagation and forgetting through an input-conditioned recurrence. Mamba-2~\cite{dao2024transformers} further develops this formulation through the state-space duality framework.

\subsection{Biologically-Inspired Liquid Networks (BI-LNs)}
In parallel, a line of research emerged within the computational neuroscience community. Hasani et al.~\cite{hasani2021liquid,hasani2018liquid} introduced Liquid-time Constant (LTC) networks as a biologically-inspired extension of CT-RNNs, replacing the fixed neuronal time constant with an input-dependent function. This allowed individual neurons to modulate their integration time scales in response to incoming signals, achieving rapid response to change when needed and slow integration during stable periods. The LTC network lies between CT-RNNs and NODEs in terms of expressiveness. It is more expressive than fixed-$\tau$ CT-RNNs but less expressive than fully general NODEs. Closed-form Continuous (CfC)~\cite{hasani2022closed,liang2026rederived} networks approximate the closed-form solution to LTC-ODE. By replacing numerical integration through a fused solver with a single forward pass, CfC achieved the same accuracy as LTC at significantly less computational cost. Both models uses Neuronal Circuit Policies (NCPs)~\cite{lechner2018neuronal}, inspired by the sparse wiring architecture of the \emph{C.\,elegans} nematode, as the core sparse connectivity mechanism to mimic the nematode's nervous system. Cantini et al.~\cite{cantini2025exact} redefined the exact closed-form formulation of LTC networks and implemented it through a computationally efficient recursive algorithm with arbitrary precision. Liquid-Resistance Liquid-Capacitance (LRCs)~\cite{farsang2024liquid} extend the LTC-ODE by introducing a state-dependent liquid capacitance and liquid resistance, improving damping of oscillations and accuracy. More recently, the Liquid Foundation Models (LFM2)~\cite{amini2025lfm2} family has scaled liquid-inspired architectures to billion parameter scale (350M--8.3B).

\subsection{Continuous-Time Transformers}
A parallel line of work extends the DT-Transformer architecture towards continuous time by incorporating temporal dynamics directly into representation learning. The multi-time attention network (mTAN)~\cite{shukla2021multi} replaced the fixed interpolation kernel of earlier methods with a learned neural attention mechanism that assigns weights to observations based on their temporal distance and context. ODEFormer~\cite{d2023odeformer} trains a transformer on synthetic trajectories to infer symbolic ODE systems directly from noisy and irregular data. ContiFormer~\cite{chen2023contiformer} replaces discrete token embedding with continuous latent trajectories governed by NODEs, enabling each element to be queried at arbitrary times. ANCDE~\cite{jhin2021ace,jhin2024attentive} integrates attention directly into the NCDE framework, producing a CT-attention signal that gates the influence of the driving path. OT-Transformer~\cite{kan2025ot} formulates entire transformer blocks as a single ODE with optimal-transport regularization. Continuous-time Attention (CTA)~\cite{chien2021continuous} embeds an attention mechanism within a NODE, allowing attention weights and hidden states to evolve jointly over time. PDE-Attention~\cite{zhang2025continuous} evolves the full attention matrix over pseudo-time via partial differential equations (PDEs).

\subsection{Hybrid Architectures}
A parallel line of research has also explored hybrid architectures that exchange traits from one branch with other branches. Liquid-S4~\cite{hasani2022liquid} integrates input-dependent dynamics of LTCs in S4 to allow the state transition to adapt continuously to the input sequence while preserving efficient parallel computation. Flexible Unified Information Dynamics (FLUID)~\cite{razzaq2026fluid} Transformer introduces the Liquid Attention Network (LAN), which derives its biological design from from LTC networks and rethinks attention itself as the solution to a linear ODE with nonlinear input-dependent gates and interlinked gates. It also replaces conventional residual connections with liquid hyper-connections~\cite{Zhu2024HyperConnections}, enabling adaptive information propagation throughout the network. Neuronal Attention Circuit (NAC)~\cite{razzaq2025neuronal} extends the LAN by approximating a closed-form solution and incorporating a repurposed NCP wiring mechanism into the gate computation to mimic the attention circuit in the \textit{C.\,elegans} nematode. Neuronal Stochastic Attention Circuit (NSAC)~\cite{razzaq2026neuronal} further extends the NAC by rethinking attention as a solution to an Ornstein–Uhlenbeck (OU)-SDE~\cite{ornsteinuhlenbeck2014overview} to jointly quantify both aleatoric and epistemic uncertainty through a two-term loss objective.

\subsection*{Key Takeaways}
The modeling of neural dynamics as a continuous process predates the deep learning revolution. Hodgkin and Huxley~\cite{hodgkin1952quantitative} formulated neural dynamics with four differential equations, and Funahashi and Nakamura~\cite{Funahashi1993} later proved that CT-RNNs are universal approximators of smooth dynamical flows. An often overlooked insight is that several discrete-sequence models are discretizations of CT systems. The Elman RNN~\cite{elman1990finding} corresponds to Euler integration of the CT-RNN, and the LSTM's forget gate~\cite{mozer2017discrete} can be interpreted as an adaptive time constant. The same mechanism was discovered in LTC networks~\cite{hasani2018liquid,hasani2021liquid,hasani2022closed} by a separate community. The modern revival of CT machine learning reformulates these classical ideas by directly parameterizing the vector field and treating time as a first-class variable rather than a discrete index.
\section{Canonical Dynamical System}~\label{section:canonical}

The canonical dynamical system governing all major branches of CT machine learning can be written as the following SDE:
\begin{equation}
dx(t) = \underbrace{f(x(t), u(t), t; \theta)}_{\text{drift}} \, dt + \overbrace{g(x(t), u(t), t; \varphi)}^{\text{diffusion}} \, d\mathcal{W}(t), \qquad y(t) = \underbrace{h(x(t), u(t); \psi)}_{\text{readout}}
\label{eq:canonical_major}    
\end{equation}
where $x(t) \in \mathbb{R}^N$ is the latent hidden state, $u(t) \in \mathbb{R}^{d_{\text{in}}}$ the input signal, $y(t) \in \mathbb{R}^{d_{\text{out}}}$ the output, $\mathcal{W}(t) \in \mathbb{R}^m$ is a Wiener process, $f$ is the deterministic drift vector field, $g$ is the diffusion function governing stochastic dynamics, and $h$ is the readout mapping. We assume $f$ and $g$ are locally \emph{Lipschitz} in $x$ and satisfy a linear growth condition in $x$, uniformly  for $t$ in compact intervals. These conditions guarantee the existence and uniqueness of a strong solution to Eqn.~\ref{eq:canonical_major}, and the stochastic integral is taken in the Itô sense. This equation can be viewed as the CT analogue of a RNN with an arbitrary nonlinear transition function, optional stochastic forcing, and a learnable readout. \\
The discrete-time counterpart is obtained by applying a discretization operator $\mathcal{D}$ with step size $\Delta_k$:
\begin{equation}
x_k = \mathcal{D}(x_{k-1}, u_k, \Delta_k; \theta)
\label{eq:major_canonical_discrete}
\end{equation}
Different CT architectures employ different discretization operators $\mathcal{D}$, leading to distinct trade-offs among numerical accuracy, stability, computational efficiency, and memory usage. The most widely adopted operators are summarized as:

\textit{\textbf{Euler discretization:}} The simplest first-order discretization approximates the derivative using the current state;
\begin{equation}
x_k = x_{k-1} + \Delta_k f(x_{k-1}, u_k; \theta)
\end{equation}
Euler integration is computationally efficient but is only first-order accurate $\mathcal{O}(\Delta_k)$, which makes it sensitive to large step sizes and stiff dynamics. Setting $\Delta_k=1$ and $f$ to the CT-RNN recovers the Elman RNN~\cite{elman1990finding}, bridging the CT-RNN and the DT-RNN literatures. 

\textit{\textbf{Zero-Order Hold (ZOH):}} The zero-order hold (ZOH) method assumes that the input remains constant over each sampling interval and integrates the linear dynamics exactly:
\begin{equation}
\bar{A}_k = e^{A(\Delta_k)},
\qquad
\bar{B}_k =
\int_{0}^{\Delta_k}
e^{A(\Delta_k-s)}B\,ds, \qquad x_k = \bar{A}_k x_{k-1} + \bar{B}_k u_k 
\end{equation}
ZOH is the standard discretization for linear SSMs. Since the matrix exponential $e^{A\Delta_k}$ exactly captures the continuous linear dynamics, the only approximation error arises from the piecewise-constant input assumption.

\textbf{\textit{Bilinear Transformation:}} The bilinear (Tustin) transform maps the CT $s$-plane to the DT $z$-plane via $s\rightarrow \frac{2}{\Delta_k}\frac{z-1}{z+1}$, yielding
\begin{equation}
\bar{A}_k =
\left(I+\frac{\Delta_k}{2}A\right)
\left(I-\frac{\Delta_k}{2}A\right)^{-1},
\qquad
\bar{B}_k =
\Delta_k\left(I-\frac{\Delta_k}{2}A\right)^{-1}B,
\end{equation}
provided that $I-\frac{\Delta_k}{2}A$ is nonsingular. Unlike Euler integration, the bilinear transform maps the open left half of the $s$-plane into the interior of the unit circle, thereby preserving stability. It is therefore often preferred when the sampling interval is relatively large compared with the system dynamics. 

\textbf{\textit{Closed-form approximation:}} For specific choices of $f$, an approximation of closed-form solution can be derived under the assumption of piecewise-constant inputs over the interval $[t, t+\Delta_k]$. CfC~\cite{hasani2022closed} and NAC~\cite{razzaq2025neuronal} exploit this property for LTC~\cite{hasani2018liquid} and LAN~\cite{razzaq2026fluid} dynamics:
\begin{equation}
x(t+\Delta_k) = (1-e^{-\Delta_k/\tau_{\mathrm{eff}}}) \odot (f(x,u)-x) +x
\end{equation}
Here, $\tau_{\mathrm{eff}}$ denotes the learned effective time constant, and $f$ is a nonlinear function of the current state and input. Closed-form approximations replace the $\mathcal{O}(S)$ computational cost of adaptive numerical solvers with $\mathcal{O}(1)$ computation per time step, at the expense of approximation errors that increase with the degree of nonlinearity in the underlying dynamics.

\textbf{\textit{Adaptive ODE solvers:}} For general nonlinear dynamics $f(x;\theta)$, the integral $
\int f(x(\tau);\theta)\,d\tau$ cannot be evaluated analytically and must instead be approximated numerically. Adaptive step-size solvers, such as \emph{Dormand--Prince (DOPRI5)} and \emph{Runge--Kutta--Fehlberg} methods, automatically adjust the integration step size to satisfy a prescribed error tolerance. Consequently, the number of function evaluations varies with the complexity and stiffness of the dynamics, ranging from only a few evaluations for smooth trajectories to hundreds or even thousands for stiff systems~\cite{wanner1996solving}. As a result, the computational cost becomes data dependent rather than fixed.

% figure 1
\begin{figure*}[t]
\centering
\includegraphics[width=0.95\textwidth]{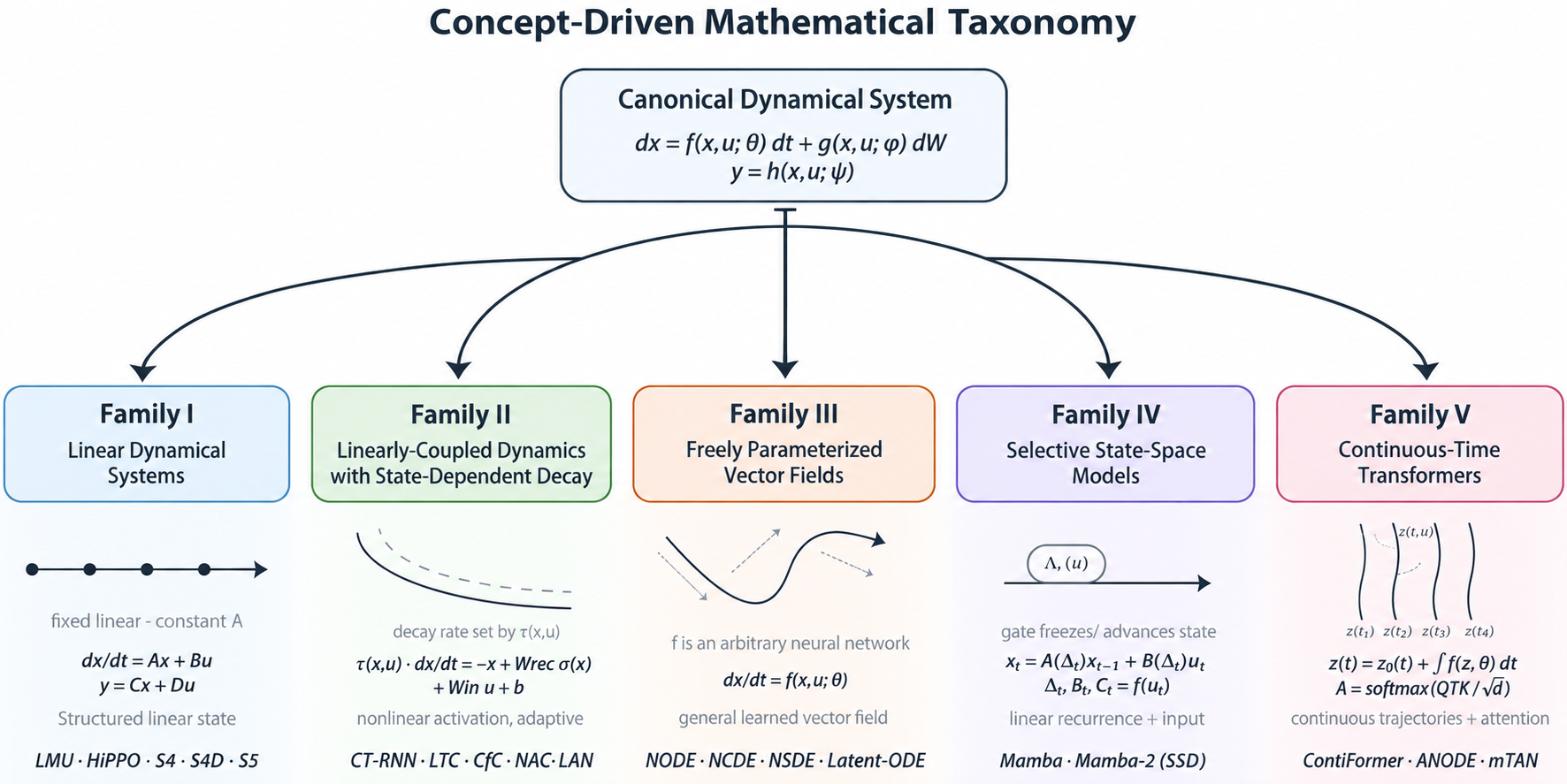}
\caption{A Visual representation of Concept-driven Taxonomy.}
\label{fig:taxonomy}
\end{figure*}

\subsection{Concept-Driven Taxonomy}

The existing surveys broadly categorize CT architectures in three taxonomic ways: (i) chronological; (ii) community-based; (iii) application-based. \emph{Chronologically}, architectures are organized according to their publication order, with each new architecture superseding its predecessor. This view is misleading as architectures are better understood as points along a continuum of design choices that has existed since the first analysis of the CT-RNNs in 1989~\cite{funahashi1989approximate}. The chronological taxonomy conflates historical order with mathematical structure, obscuring the algebraic relationship among seemingly diverse architectures. In the \emph{ community-based} taxonomy the models are grouped by their community of origin. Under this, CT-RNNs are associated with computational neuroscience, S4 with signal processing, and NDEs with deep learning. This reflects how the field grew, but it does not capture the underlying mathematical connections between branches. \emph{In application-based} the architectures are grouped by their application domain such as time-series forecasting, sequence modeling, generative modeling. It conflates architectural design with downstream tasks and makes it difficult to identify shared principles or transfer insight across branches.

We introduce a taxonomy that organizes all branches by two axes: (i) the constraint imposed on the drift $f$ (linear $\to$ partially structured $\to$ freely parametrized) that determines how the hidden state evolves in continuous time; and (ii) the memory/interaction mechanism (single recurrent state, selective scan, pairwise attention). Viewed through this lens, models developed across different research communities often share common dynamical principles even when their terminology differs. Branches of CT machine learning fall into five families: (i) \textit{Linear Dynamical Systems}; (ii) \textit{Linearly-Coupled Dynamics with State-Dependent Decay}; (iii) \textit{Freely Parameterized Vector Fields}; (iv) \textit{Selective Scan State-Space Models}; and (v) \textit{Continuous-Time Transformers}. Table~\ref{tab:ct_taxonomy} summarizes families according to their functional form with representative architectures. The taxonomy captures one central design decision: how much structure is imposed on $f$, and the resulting trade-off between expressiveness and computational efficiency. Figure~\ref{fig:taxonomy} illustrates the taxonomy tree of families corresponding to their base functional form and canonical dynamical system.

\begin{table*}[t]
\centering
\caption{Taxonomy of CT machine learning models based on their underlying functional formulations.}
\label{tab:ct_taxonomy}
\resizebox{0.95\textwidth}{!}{
\begin{tabular}{c m{3.5cm} m{3cm} m{3.2cm} m{3cm} m{4.5cm}}
\toprule
\textbf{Family} & \textbf{Canonical dynamics} & \textbf{Nonlinearity source} &
\textbf{Parameter adaptivity} & \textbf{Memory mechanism} &
\textbf{Representative models} \\
\midrule
\textbf{I} &
$\dot{x}=Ax(t)+Bu(t)$ &
None (linear dynamics) &
None (time-invariant) &
Linear recurrence / convolution kernel &
LMU~\cite{voelker2019legendre}, HiPPO~\cite{gu2020hippo}, SHiPPO~\cite{mizuguchi2026shippo}, DeepState~\cite{rangapuram2018deep}, S4~\cite{gu2021efficiently}, LSSL~\cite{gu2021combining}, LRU~\cite{orvieto2023resurrecting}, S4D~\cite{gu2022parameterization}, DSS~\cite{gupta2022diagonal}, S5~\cite{smith2208simplified}, GSS~\cite{mehta2022long}, H3~\cite{dao2022hungry}, NCDSSM~\cite{ansari2023neural}, BiGS~\cite{wang2022pretraining},  S4ND~\cite{nguyen2022s4nd}, SaShiMi~\cite{goel2022sashimi} \\
\midrule
\textbf{II} &
$\tau(x,u)\odot\dot{x}=-x(t)+W_{rec}\sigma(x(t))+W_{in}u(t)+b$ &
Fixed nonlinear activation or analytical gating &
State- or input-dependent time constants / gates &
Continuous hidden state &
CT-RNNs~\cite{funahashi1989approximate}, CT-GRU~\cite{debrouwer2019gru,mozer2017discrete}, PhasedLSTM~\cite{neil2016phased}, LTC~\cite{hasani2018liquid, hasani2022liquid}, CfC~\cite{hasani2022closed, liang2026rederived, cantini2025exact}, LAN~\cite{razzaq2026fluid}, FLUID~\cite{razzaq2026fluid}, Liquid-S4~\cite{hasani2022liquid}, NAC~\cite{razzaq2025neuronal}, NCPs~\cite{lechner2018neuronal}, NSAC~\cite{razzaq2026neuronal}, coRNN~\cite{rusch2021cornn}, cPLRNNs~\cite{brandle2026continuous}\\
\midrule
\textbf{III} &
$\dot{x}=f(x(t),u(t),t;\theta)$ &
Arbitrary neural vector field &
Learned implicitly through $f$ &
Continuous hidden state or controlled trajectory &
NODE~\cite{chen2018neural}, NCDE~\cite{kidger2020neural}, NSDE~\cite{tzen2019neural, oh2024stable}, NRDE~\cite{morrill2021neuralroughdifferentialequations,lyons2007differential}, Latent-ODE~\cite{rubanova2019latent},  ODE-LSTM~\cite{lechner2022mixed}, ANODEs~\cite{dupont2019augmented,ijcai2019p103}, SONODEs~\cite{bodnar2020second}, NJSDEs~\cite{jia2019neural}, LSDE~\cite{oh2024stable}, GSDE~\cite{mathieu2023geometric}, FIM-ODE~\cite{mauel2026foundation}, SINODE~\cite{zhang2025semi}, PNODE~\cite{lee2021parameterized}, SCOTCH~\cite{wang2024neural}\\
\midrule
\textbf{IV} &
$x_k=\bar{A}_k(\Delta_k)x_{k-1}+\bar{B}_k(\Delta_k)u_{k},\quad y=C_kx_k$ &
Input-conditioned state-space parameters &
$\bar{A}_k$, $\bar{B}_k$, and adaptive discretization $\Delta_k$ &
Selective recurrent scan &
Mamba~\cite{gu2023mamba}, Mamba-2~\cite{dao2024transformers}, Jamba~\cite{lieber2024jamba},  Vim~\cite{zhu2024vision}, VMamba~\cite{liu2024vmamba}, MambaByte~\cite{wang2024mambabyte}, MambaML~\cite{zhu2025mambaml}, CMamba~\cite{zhang2024cmamba}, MambaTS~\cite{cai2024mambatsimprovedselectivestate},  RetNet~\cite{sun2023retentive}, RWKV~\cite{peng2023rwkv}, Griffin~\cite{de2024griffin}, GLA~\cite{yang2023gated}, S7~\cite{soydan2024s7selectivesimplifiedstate}, MuonSSM~\cite{nguyen2026muonssm} \\
\midrule
\textbf{V} &
$\dot{z}=f(z),\quad
\alpha_{ij}=\operatorname{Attn}(z_i,z_j,\Delta_k )$ &
Neural dynamics combined with attention &
Content- and time-dependent attention weights &
Attention over continuous trajectories &
mTAN~\cite{shukla2021multi}, ANCDE~\cite{jhin2021ace,jhin2024attentive}, CTA~\cite{chien2021continuous}, ContiFormer~\cite{chen2023contiformer}, OT-Transformer~\cite{kan2025ot}, PDE-Attention~\cite{zhang2025continuous}, WrapFlow~\cite{shen2026enhancing}, CTLPE~\cite{kim2024continuous}.\\
\bottomrule
\end{tabular}
}
\end{table*}

\subsubsection*{Family I: Linear Dynamical Systems}

Linear dynamical systems are the classical CT framework for modeling temporal evolution, representing hidden-state dynamics and observations through classical linear state-space equations:
\begin{equation}
\frac{dx}{dt} = A x(t) + B u(t), \qquad y(t) = C x(t) + D u(t),
\end{equation}
with $A \in \mathbb{R}^{N \times N}$, $B \in \mathbb{R}^{N \times d_{\text{in}}}$, $C \in \mathbb{R}^{d_{\text{out}} \times N}$, $D \in \mathbb{R}^{d_{\text{out}} \times d_{\text{in}}}$. The state evolution is strictly linear in $x(t)$ and $u(t)$, with no nonlinear activation or stochastic component. This formulation scheme has been studied extensively in control theory and has recently attracted interest in machine learning through the HiPPO framework~\cite{gu2020hippo}, which derives $A$ from an optimal projection of the input history and thereby structures it for long-term memory. Earlier linear SSMs with a learned $A$ suffered from vanishing gradients. Moreover the eigenvalues of $\bar{A} = e^{A\Delta_k}$ can drift away from the unit circle during training, and the resulting recurrent state grows or decays exponentially.

\textbf{\textit{Structural variants:}} The key design choice is how to parameterize and constrain the transition matrix imposed on $A$:
\begin{itemize}
    \item[-] \textit{Untrained structured $A$:} LMU~\cite{voelker2019legendre} derives the state matrix $A$ from Legendre polynomials, implementing an optimal projection of the input history onto a polynomial basis over a sliding window. Both $A$ and $B$ are fixed analytically, while only the output matrices $C$ and $D$ are learned.
    \item[-] \textit{HiPPO-initialized $A$:} HiPPO~\cite{gu2020hippo} generalizes the LMU by formalizing online function approximation. Given a measure $\mu(t,\vartheta)$ that weights past observations, the optimal $N$-dimensional compression of the input history into a polynomial basis evolves according to $\dot{x}=A(t)x+B(t)u$, where $A(t)$ and $B(t)$ are derived analytically from the chosen measure. \textit{LegS} variant assigns equal weight to the entire history under a scaled Legendre measure, providing unbounded long-range memory; \textit{LagT} implements exponentially decaying memory with a controllable time constant; and \textit{LegT} provides bounded finite-memory representations.
    \item[-] \textit{Diagonal-plus-low-rank (DPLR):} S4~\cite{gu2021efficiently} parameterizes the HiPPO state matrix $A$ using DPLR representation, enabling efficient computation of the convolution kernel $K_i = C\bar{A}^i\bar{B}$ through the Cauchy kernel. The HiPPO-derived initialization preserves the desired memory properties while allowing the DPLR parameters to be optimized during training.
    \item[-] \textit{Diagonal $A$:} S4D~\cite{gu2022parameterization} and DSS~\cite{gupta2022diagonal} eliminate the low-rank correction of DPLR and directly parameterize $A$ as a diagonal complex matrix.
    \item[-] \textit{Multi-head SSMs:} Inspired by the multi-head attention mechanism~\cite{vaswani2017attention}, S5~\cite{smith2208simplified} introduces multiple independent SSMs processed in parallel with a diagonal $A$ shared across heads.
\end{itemize}

\textbf{\textit{Representative architectures:}} LMU~\cite{voelker2019legendre}, HiPPO~\cite{gu2020hippo}, DeepState~\cite{rangapuram2018deep}, SHiPPO~\cite{mizuguchi2026shippo}, S4~\cite{gu2021efficiently}, LSSL~\cite{gu2021combining}, LRU~\cite{orvieto2023resurrecting}, S4D~\cite{gu2022parameterization}, DSS~\cite{gupta2022diagonal}, S5~\cite{smith2208simplified}, GSS~\cite{mehta2022long}, H3~\cite{dao2022hungry}, BiGS~\cite{wang2022pretraining}, S4ND~\cite{nguyen2022s4nd}, NCDSSM~\cite{ansari2023neural}, SaShiMi~\cite{goel2022sashimi}.

\textbf{\textit{Key takeaways:}} Three key points emerge from this family. \textit{First}, the parameterization of $A$ governs long-range memory: LMU and HiPPO show that carefully structured linear dynamics can retain historical information over long timescales without needing nonlinear state transitions. \textit{Second}, the convolution view splits training and inference into complementary regimes. Training runs as a linear-complexity parallel convolution, while inference uses a constant-time recurrent update. \textit{Third}, diagonalization is important: the progression from DPLR to diagonal to scalar-time-identity parameterization suggests that the expressive power of SSMs does not require dense $A$, and a diagonal matrix $A$ with $N$ learned complex eigenvalues can represent $N$ frequencies or decay rates.

\subsubsection*{Family II: Linearly-Coupled Dynamics with State-Dependent Decay}

State-dependent dynamical systems describe CT evolution in which the rate of state relaxation is modulated by the current hidden state and the external input. Each latent dimension can adapt its temporal behavior to the evolving dynamics, which gives a mechanism for representing heterogeneous, context-dependent memory. The base equation is
\begin{equation}
\tau(x, u) \odot \frac{dx}{dt} = -x(t) + W_{\text{rec}} \, \sigma(x(t)) + W_{\text{in}} \, u(t) + b,
\end{equation}
where $\sigma$ is a nonlinear \emph{sigmoid} or \emph{tanh} activation, $\tau(x,u)$ is a vector of time constants, and $\odot$ is element-wise multiplication. The vector field splits into a \emph{linear decay} component ($-x$) and a \emph{nonlinear interaction} component ($W_{\text{rec}} \,\sigma(x)$), with a time constant controlling their relative contribution. This partial structure is what distinguishes \emph{Family II} from the fully linear \emph{Family I} and the unconstrained \emph{Family III}.

\textbf{\textit{Structural variants:}} The key distinguishing factor of this family is the complexity of $\tau(x,u)$:
\begin{itemize}
\item[-] \textit{CT-RNNs}~\cite{funahashi1989approximate}: $\tau$ is a scalar constant.
\item[-] \textit{CT-GRU}~\cite{debrouwer2019gru,mozer2017discrete}: The CT limit of the GRU provides the vector field; a closed-form approximate solution eliminates the ODE solver; and a Bayesian observation update provides calibrated uncertainty estimates.
\item[-] \textit{LTC}~\cite{hasani2018liquid}: $\tau_i(x, u) = \tau_{\text{base},i} + w_{\tau,i} \, \sigma([x(t), u(t)]; \theta_{\tau,i})$ is a learned function of state and input. The LTC can be rewritten as $\frac{dx}{dt} = -\operatorname{diag}\!\left(\tau^{-1}\right)x(t) + f\!\left(x(t),u(t);\theta\right)$, making explicit that it replaces the scalar $\tau$ with an input-dependent diagonal matrix which controls the decay rate per dimension. 
\item[-] \textit{LAN}~\cite{razzaq2026fluid}: LAN borrows the idea of input-dependent time-constants from LTC networks and applies to attention-mechanisms:  $\frac{da}{dt} = -\text{diag}(\tau^{-1} \cdot a(t) + f(x(t), u(t); \theta)$. Instead of modeling states, the LAN-ODE models attention logits, which evolve as the solution to the ODE.
\item[-] \textit{CfC}~\cite{hasani2022closed}: LTC-ODE is replaced by an approximate closed-form solution: $x(t+\Delta_k) = (1 - e^{-\Delta_k/\tau_{\text{eff}}}) \odot (f(x,u;\theta) - x) + x$, where $\tau_{\text{eff}}$ is a learned effective time constant. This replaces $S$ ODE-solver evaluations per interval with a single closed-form computation, reducing training and inference time.
\item[-] \textit{NCPs}~\cite{lechner2018neuronal}: The recurrent weights matrix $W_{\text{rec}}$ is constrained to match the sparse connectivity pattern of the \emph{C.\, elegans} nematode, where each neuron receives input from approximately 12-15 presynaptic partners~\cite{White1986, Cook2019}.
\item[-] \textit{Liquid-S4}~\cite{hasani2022liquid}: Introduces input-dependent liquid dynamics into S4 to enable adaptive dynamics.
\item[-] \textit{NAC}~\cite{razzaq2025neuronal}: NAC approximates LAN by approximating its closed-form solution with frozen coefficient constants and modeling gates using a repurposed NCP wiring to mimic the attention circuit of \emph{C.\, elegans}. To address the quadratic computational complexity of LAN, NAC implements a partitioning-based query-key interactions, achieving $\mathcal{O}(n\sqrt{n}k)$.
\item[-] \textit{NSAC}~\cite{razzaq2026neuronal}: NSAC generalizes NAC by modeling attention logits with a closed-form approximation of a gated Ornstein-Uhlenbeck SDE, where gates are modulated using a repurposed NCP wiring mechanism. It imposes a Gaussian distribution over the attention logits and produces uncertainty estimates at the representational level.
\end{itemize}
A useful dynamical interpretation, usually obscured from the literature, is that the LSTM cell-state update can be viewed as a discrete-time approximation to a CT recurrent system with input-dependent effective time constants. The cell-state dynamics $c_k = f_k \odot c_{k-1} + i_k \odot \tilde{c}_k$ are algebraically equivalent to an Euler discretization of a leaky CT-RNN~\cite{mozer2017discrete}, in which the forget gate  $f_k$ governs the effective decay rate (the effective time constant $\Delta_k / (1-f_k)$) while the input gate $i_k$ regulates new information. LSTM and LTC independently converged on the same mechanism of adaptive time constants through different intellectual traditions.

\textbf{\textit{Representative architectures:}} CT-RNNs~\cite{funahashi1989approximate}, PhasedLSTM~\cite{neil2016phased},  CT-GRU~\cite{debrouwer2019gru,mozer2017discrete}, LTC~\cite{hasani2018liquid, hasani2022liquid}, CfC~\cite{hasani2022closed, liang2026rederived, cantini2025exact}, FLUID/LAN~\cite{razzaq2026fluid},  NCPs~\cite{lechner2018neuronal}, Liquid-S4~\cite{hasani2022liquid}, NAC~\cite{razzaq2025neuronal}, NSAC~\cite{razzaq2026neuronal}, coRNN~\cite{rusch2021cornn}, and cPLRNNs~\cite{brandle2026continuous}.

\textbf{\textit{Connections to other families:}} The CT-RNN reduces to a linear SSM (\emph{Family I}) when $\sigma$ is the identity and $\tau = 1$, giving $\frac{dx}{dt} = (W_{\text{rec}} - I)x + W_{\text{in}} u = Ax + Bu$. It becomes a NODE (\emph{Family III}) when its specific functional form is absorbed into a general $f(x, u; \theta)$. \emph{Family II} is the intersection of structured linear dynamics and unconstrained neural dynamics, and is more expressive than the former and more computationally efficient than the latter.

\textbf{\textit{Key takeaways:}} Three key points are observed from this family. \textit{First}, by explicitly parameterizing the integration time constant $\tau(x,u)$, LTC and LAN separate temporal memory from nonlinear state transformation, making the resulting dynamics more interpretable, analyzable, and independently controllable. \textit{Second}, CfC and NAC show that the cost of numerical integration is the main barrier to deploying these models at scale, and the closed-form solution addresses it. \textit{Third}, biological constraints trade capacity for sample efficiency: NCPs and NAC are robust to distribution shift~\cite{lechner2020neural, razzaq2023neural, razzaq2024neural}, but they do not improve accuracy on large-scale benchmarks where data are abundant.

\subsubsection*{Family III: Freely Parameterized Vector Fields}

Freely parameterized vector fields describe CT dynamics in which the governing differential equation is a learnable function $f$ with no predefined analytical structure. In contrast to \emph{Family I \& II}, these models learn the vector field itself from data, so the dynamics can approximate a broad class of nonlinear CT systems. The base equation is
\begin{equation}
\frac{dx}{dt} = f(x(t), u(t), t; \theta), \qquad x(t_L) = x(t_0) + \int_{t_0}^{t_L} f(x(s), u(s); \theta) \, ds,
\end{equation}
where $f$ is a neural network. Because no structural constraint is imposed on $f$, the expressive power of the model depends primarily on the chosen architecture, and the vector field is learned directly from data. Practitioners can design $f$ for the problem at hand rather than adapt the problem to a pre-specified $f$. The resulting trajectories are obtained by numerical integration, so the computational cost depends on the number of solver evaluations rather than on a single recurrent update.

\textbf{\textit{Structural variants:}} The members of this family differ primarily in the information available to the vector field $f$ and the mathematical formulation of differential equation:
\begin{itemize}
\item[-] \textit{NODE}~\cite{chen2018neural}: $f$ is a multilayer perceptron. The solution is obtained by calling an adaptive-step ODE solver, and gradients are computed via the adjoint sensitivity method~\cite{Pontryagin1962Mathematical} or by backpropagating through the solver with checkpointing. 
\item[-] \textit{NCDE}~\cite{kidger2020neural}: Replaces the time derivative with a Riemann-Stieltjes integral: $x(t) = x(t_0) + \int_{t_0}^{t} f(x(s); \theta) \, dU(s)$, where $U(s)$ is a continuous path interpolating the observations. The differential $dU(s)$ encodes the variation of the input rather than its value at isolated points. The path formulation makes the solution continuous in the input path, so the model is stable to input perturbation and respects the data's causal continuity.
\item[-] \textit{NRDE}~\cite{morrill2021neuralroughdifferentialequations, lyons2007differential}: Extends the CDE to rough paths, handling inputs that are not differentiable or off-bounded variation.
\item[-] \textit{Latent-ODE}~\cite{rubanova2019latent}: An encoder-ODE-decoder architecture. The ODE-RNN variant hybridizes this with discrete updates at observation times, providing a direct bridge between the fully discrete RNN and the fully continuous NODE.
\item[-] \textit{NSDE}~\cite{tzen2019neural, li2020scalable, kong2020sde}: Adds a diffusion term: $dx(t) = f(x, u; \theta) dt + g(x, u; \varphi) d\mathcal{W}(t)$. The diffusion serves two purposes: (i) as a generative model, it defines a distribution over trajectories; (ii) as a regularizer, it prevents the deterministic drift from overfitting to observational noise. This diffusion mechanism is qualitatively different from weight decay or dropout because it operates at the trajectory level rather than the parameter level.
\end{itemize}

\textbf{\textit{Representative architectures:}} NODE~\cite{chen2018neural}, NCDE~\cite{kidger2020neural}, NRDE~\cite{morrill2021neuralroughdifferentialequations,lyons2007differential}, Latent-ODE~\cite{rubanova2019latent}, ODE-LSTM~\cite{lechner2022mixed}, ANODEs~\cite{dupont2019augmented}, SONODEs~\cite{bodnar2020second}, NJSDEs~\cite{jia2019neural}, HNN~\cite{greydanus2019hamiltonian}, LNN~\cite{cranmer2020lagrangian}, Graph ODE~\cite{poli2021graph}, LSDE~\cite{oh2024stable}, GSDE~\cite{mathieu2023geometric}, FIM-ODE~\cite{mauel2026foundation}, SINODE~\cite{zhang2025semi}, PNODE~\cite{lee2021parameterized}, SCOTCH~\cite{wang2024neural}.

\textbf{\textit{Key takeaways:}} Three important points are noted from this family. \textit{First}, increasing dynamical flexibility introduces computational trade-offs. Although the adjoint sensitivity method reduces memory requirements by avoiding storage of intermediate solver states, gains are narrower than originally claimed. Gholaminejad et al.~\cite{ijcai2019p103} demonstrated that discretize-then-optimize is often faster and more accurate than the adjoint method. Rackauckas et al.~\cite{rackauckas2020universal} concluded that the discretize-then-optimize approach with checkpointing offers the best memory-accuracy trade-off for most problems. \textit{Second}, the path-based formulation of NCDEs is more natural for irregular sampling than the interpolate-then-integrate approach of ODEs~\cite{kidger2020neural, jhin2024attentive}. \textit{Third}, stochasticity is a regularizer, not merely a modeling choice: it prevents overfitting in ways that deterministic regularizers do not~\cite{bilos2021neuralflows,kidger2022neural}, effectively providing data augmentation in function space.

\subsubsection*{Family IV: Selective Scan State-Space Models}

Input-selective SSMs formulate sequence dynamics through a linear state transition whose parameters are modulated by the input. Instead of learning a fixed transition operator shared across all time steps, these models let the discretization parameters and input/output projections vary with the current observations, which yields an adaptive dynamical system:
\begin{equation}
x_k = \bar A_k(\Delta_k)x_{k-1}+\bar B_k(\Delta_k,u_k)u_k,\ y_k=C_k(u_k)x_k,
\end{equation}
where $\Delta_k = \text{softplus}(W_{\Delta_k} u_k + b_{\Delta_k})$, and $\bar{A}_k(\Delta_k), \bar{B}_k(\Delta_k)$ are obtained by zero-order-hold (ZOH)~\cite{pechlivanidou2022zero} discretization of a continuous linear system with diagonal $A$ evaluated at the input-dependent step size. The state transition remains linear, but the system matrices become input-dependent, so the model defines a time-varying linear system. This family occupies a distinctive position in the taxonomy: it inherits the computational efficiency of \emph{Family I} while acquiring the adaptivity of \emph{Family II \& III}. Its defining architectural feature is the selection mechanism, which makes $\Delta_k$, $B$, and $C$ functions of the input.

\textbf{\textit{Structural variants:}} The key design distinction is the input-selective and adaptive mechanism:
\begin{itemize}
\item[-] \textit{Mamba}~\cite{gu2023mamba}: The selection mechanism serves a dual role. \textit{First}, it is a gating mechanism, where small $\Delta_k$ freezes the state, equivalent to a forget gate of $1$, ignoring irrelevant inputs; large $\Delta_k$ rapidly updates the state, incorporating informative inputs. \textit{Second}, it is a selection mechanism in the information-theoretic sense; the model chooses to process or ignore each token based on its relevance to the task.
\item[-] \textit{Mamba-2/SSD}~\cite{dao2024transformers}: Further constrains $A$ to a scalar times identity ($A = aI$), revealing that the SSM recurrence is structurally dual to a particular form of linear attention: $Y = (L \circ \text{causal}(Q K^\mathsf{T})) V$ , where $L$ is a structured multiplicative mask determined by $a$. This duality enables Mamba-2 to achieve faster training and linear inference.
\item[-] \textit{Jamba}~\cite{lieber2024jamba}: Interleaves Mamba selective state-space layers with attention layers and mixture-of-experts (MoE) components.
\item[-] \textit{Vim}~\cite{zhu2024vision}: Adapts selective SSMs for visual representation learning by applying bidirectional scanning mechanisms to image patch sequences.
\end{itemize}

\textbf{\textit{Representative architectures:}} The family includes Mamba~\cite{gu2023mamba}, Mamba-2~\cite{dao2024transformers}, Jamba~\cite{lieber2024jamba}, Vim~\cite{zhu2024vision}, VMamba~\cite{liu2024vmamba}, MambaByte~\cite{wang2024mambabyte}, MambaML~\cite{zhu2025mambaml}, Samba~\cite{ren2024samba}, CMamba~\cite{zhang2024cmamba}, MambaTS~\cite{cai2024mambatsimprovedselectivestate}, RetNet~\cite{sun2023retentive}, RWKV~\cite{peng2023rwkv}, Griffin~\cite{de2024griffin}, GLA~\cite{yang2023gated}, S7~\cite{soydan2024s7selectivesimplifiedstate}, and MuonSSM~\cite{nguyen2026muonssm}.

\textbf{\textit{Why Family IV is distinct from \emph{Family I \& II}:}} Compared with \emph{Family I}, the critical difference is the loss of the convolution model when $B_k$ and $C_k$ vary per time step: the kernel can no longer be precomputed, and Mamba recovers efficiency through the parallel associative scan rather than through convolution. Memory is \emph{content-aware} rather than a fixed convolution kernel. Compared with \emph{Family II}, the distinction lies in the gating mechanism. \emph{Family II} gates through multiplicative interaction between continuous activations, whereas the selection mechanism gates through input-dependent discretization of a linear system, which is simpler and more efficient.

\textbf{\textit{Key takeaways:}} Three important points emerge from this family. \textit{First}, the selective scan in Mamba was a significant architectural change that delivered large performance gains over S4. \textit{Second}, linear dynamics with input-dependent discretization can match the performance of nonlinear dynamics with fixed discretization. Mamba achieved transformer-competitive performance using linear state transitions, which suggests that the nonlinearity required for sequence modeling can be offloaded from the state update to input-dependent parameterization. \textit{Third}, SSMs and attention may be more closely related than their formulations suggest. The state-space duality (SSD) result~\cite{dao2024transformers,katharopoulos2020transformers} is strong evidence of a close relationship between recurrent and attention-based architectures. Whether this duality extends beyond the scalar-$A$ setting to the more general matrix-valued case remains an open theoretical question.

\subsubsection*{Family V: Continuous-Time Transformers}

CT-Transformers represent a sequence as a collection of continuously evolving latent trajectories that interact through attention computed at arbitrary time points. Rather than maintaining a single shared hidden state, each sequence element follows its own CT dynamics and exchanges information with the other elements through time-aware attention. This formulation accommodates irregularly sampled observations and CT queries. The base functional forms are
\begin{equation}
    z_i(t_q) = z_i(t_i) + \int_{t_i}^{t_q} f(z_i(s); \theta) \, ds, \qquad \alpha(t_q, t_k) = \text{softmax}\left(\frac{Q(t_q)^\mathsf{T} K(t_k)}{\sqrt{d}}\right), \qquad Q(t_q) = W_Q z_q(t_q).
\end{equation}
The state consists of a set of continuous trajectories $\{z_i(t)\}$, one per sequence element. Interaction between elements is modulated by attention over continuous time. \emph{Family V} therefore encodes temporal structure in the pairwise interactions of continuously evolving representations rather than in the evolution of a shared hidden state. Unlike conventional self-attention, which relies on discrete positional indices, CT-Transformers represent the temporal relation explicitly as a function of time ($t_q - t_k$), which supports reasoning over irregularly sampled observations and arbitrary query times.

\textbf{\textit{Structural variants:}} The members of this family differ primarily in how continuous trajectories are represented and how temporal information is incorporated into the attention mechanism:
\begin{itemize}
\item[-] \textit{mTAN}~\cite{shukla2021multi}: The simplest form: $\text{mTAN}(t_q) = \sum_k \alpha(t_q, t_k) u_k$, where $\alpha(t_q, t_k) = \text{softmax}(\text{MLP}(\kappa(t_q - t_k), \gamma(t_q), \gamma(t_k)))$ is a learned function of the time interval and absolute times. The attention can be asymmetric, context-dependent only on times, with context-dependence added through a separate mechanism.
\item[-] \textit{ANCDE}~\cite{jhin2021ace,jhin2024attentive}: Integrates attention as a dynamical system through dual coupled NCDEs:$\frac{d\alpha}{dt} = f_\alpha(\alpha; \theta_\alpha) dU(t)$, $\frac{dx}{dt} = f_x(x, \alpha; \theta_x) dU(t)$. The attention signal $\alpha(t)$ is itself a continuous function of time, co-evolving with the state. When $\alpha(t) \approx 0$, the input’s influence is suppressed; when  $\alpha(t) \approx 1$, it is fully transmitted.
\item[-] CTA~\cite{chien2021continuous}: Embeds attention mechanism within an NODE, allowing attention weights and hidden states to evolve jointly over time.
\item[-] \textit{ContiFormer}~\cite{chen2023contiformer}: Each element is a continuous trajectory $z_i(t)$ generated by a NODE from its observation time $t_i$ to the query time $t_q$. The $Q,K,V$ projections are applied at query times, producing $Q(t_q), K(t_k), V(t_k)$. The attention score depends upon the actual time interval through trajectory evolution: two elements at the same ordinal position but with different time gaps produce different attention weights.
\item[-] \textit{OT-Transformer}~\cite{kan2025ot}: Formulates entire transformer blocks as a single ODE with optimal-transport regularization. 
\item[-] \textit{PDE-Attention}~\cite{zhang2025continuous}: Evolves the full attention matrix over pseudo-time via PDEs. 
\end{itemize}

\textbf{\textit{Representative architectures:}} The family includes mTAN~\cite{shukla2021multi}, ANCDE~\cite{jhin2021ace,jhin2024attentive}, CTA~\cite{chien2021continuous}, ContiFormer~\cite{chen2023contiformer}, OT-Transformer~\cite{kan2025ot}, PDE-Attention~\cite{zhang2025continuous}, WrapFlow~\cite{shen2026enhancing}, CTLPE~\cite{kim2024continuous}.

\textbf{\textit{Key takeaways:}} CT-Transformers are the least mature family, and their quadratic attention complexity restricts most applications to shorter sequences. The conceptual boundary between CT-Transformers and recurrent models is nonetheless less distinct than it first appears: SSM kernels, selective scan operators, and explicit CT-Transformers all implement a time-dependent weighted aggregation of historical observations. They differ mainly in whether the weighting function is conditioned on input content, and in whether the aggregation happens through recurrent state updates or explicit pairwise attention computations.

\subsection{How Each Family Instantiates the Canonical Form}

Each family can be interpreted as a special case of Eqn.~\ref{eq:canonical_major} by imposing different structural constraints on $f$, $g$, and $h$. Setting $g \equiv 0$ recovers a deterministic ODE, choosing $f$ to be linear in $x$ yields a linear SSM, and parameterizing $f$ with an arbitrary neural network gives a NODE. Table~\ref{tab:ct_model_comparison} provides an overview of each architecture and the corresponding design choices within the canonical mathematical form.

\section{Training, Optimization, and Trade-offs}\label{section:training}

\subsection{Training Algorithms}
Training approaches for CT models are generally categorized into three algorithms with distinct computational and memory trade-offs: (i) backpropagation through time (BPTT); (ii) adjoint sensitivity methods; and (iii) parallel associative scan.

\begin{landscape}
\begin{table}[t]
\centering
\vspace{-17mm}
\caption{Initialization of representative architectures of each family from canonical mathematical formulation.}
\vspace{1.0mm}
\label{tab:ct_model_comparison}
\resizebox{1.4\textwidth}{!}{%
\begin{tabular}{cccccc}
\toprule
\textbf{Family / Model} &
\textbf{$f(x,u,t;\theta)$ in Continuous Time} &
\textbf{Diffusion $g$} &
\textbf{$\mathcal{D}$ (Discretization)} &
\textbf{Observation Handling} &
\textbf{Distinguishing Constraint} \\
\midrule

\multicolumn{6}{c}{\textbf{Canonical Mathematical Form}}\\
\midrule
Canonical Form &
$\frac{dx}{dt}=f(x,u,t;\theta)$ &
$g(x,u;\varphi)d\mathcal{W}(t)$ &
$\mathcal{D}(x_{k-1},u_k,\Delta_k)$ &
$y(t)=h(x(t),u(t);\psi)$ &
$\times$ \\
\midrule
\multicolumn{6}{c}{\textbf{Family I: Linear Dynamical Systems}}\\
\midrule
LMU \cite{voelker2019legendre} &
$\dot{x}=Ax+Bu$, $A$ from Legendre &
$\times$ &
ZOH, fixed $\Delta_k$ &
Convolution $y=K*u$ &
Fixed $A,B$; only $C,D$ learned \\
HiPPO \cite{gu2020hippo} &
$\dot{x}=A(t)x+B(t)u$ &
$\times$ &
ZOH / bilinear &
Convolution / recurrence &
Optimal history compression \\
S4 \cite{gu2021efficiently} &
$\dot{x}=Ax+Bu$, HiPPO init &
$\times$ &
ZOH / bilinear &
Convolution (Cauchy kernel) &
DPLR parameterization of $A$ \\
LSSL \cite{gu2021combining} &
$\dot{x}=Ax+Bu$, HiPPO init &
$\times$ &
ZOH / bilinear &
Convolution (polynomial filters) &
Unifies state-space and convolution views \\
S4D / DSS \cite{gu2022parameterization,gupta2022diagonal} &
$\dot{x}=Ax+Bu$, diagonal $A$ &
$\times$ &
ZOH &
Convolution (diagonal) &
Complex diagonal $A$ \\
LRU \cite{orvieto2023resurrecting} &
$\dot{x}=Ax+Bu$, complex diagonal $A$ &
$\times$ &
Exponential, $\bar{A}=\mathrm{diag}(e^{-\Delta_k\odot\nu})$ &
Recurrence &
Stable exponential parameterization \\
S5 \cite{smith2208simplified} &
$\dot{x}_h=Ax_h+B_hu$ &
$\times$ &
ZOH &
Parallel scan &
Multi-head diagonal SSM \\
GSS \cite{mehta2022long} &
$\dot{x}=Ax+Bu$, diagonal $A$ &
$\times$ &
ZOH &
Gated recurrence / scan &
Gated SSM for long sequences \\
H3 \cite{dao2022hungry} &
$\dot{x}=Ax+Bu$ (shift-SSM) &
$\times$ &
ZOH &
SSM + attention hybrid &
SSM replaces attention for long context \\
BiGS \cite{wang2022pretraining} &
$\dot{x}=Ax+Bu$ (bidirectional) &
$\times$ &
ZOH &
Bidirectional scan &
Bidirectional SSM for pretraining \\
S4ND \cite{nguyen2022s4nd} &
$\dot{x}=Ax+Bu$ ($N$-dim, separable) &
$\times$ &
ZOH (multidimensional) &
Multidimensional convolution &
$N$-dimensional state space \\
\midrule

\multicolumn{6}{c}{\textbf{Family II: Linearly-Coupled Dynamics with State-Dependent Decay}}\\
\midrule
CT-RNN \cite{funahashi1989approximate} &
$\dot{x}=-x+W_{\mathrm{rec}}\sigma(x)+W_{\mathrm{in}}u+b$ &
$\times$ &
Euler, $\Delta_k$ from data &
Recurrence &
Fixed $\tau$, sigmoidal $\sigma$ \\
PhasedLSTM \cite{neil2016phased} &
Standard LSTM with time gate $\phi(t)$ &
$\times$ &
Euler, $\Delta_k=1$ with $\phi(t)$ &
Discrete recurrence gated by $\phi(t)$ &
Periodic continuous-time gate \\
CT-GRU~\cite{debrouwer2019gru,mozer2017discrete} &
$\dot{x}=\mathrm{GRU}_{\mathrm{ct}}(x,u,t;\theta)$ &
$\times$ &
Closed-form (GRU-based update) &
Recurrence with continuous gating &
GRU gates in continuous time \\
LTC \cite{hasani2021liquid} &
$\dot{x}=-\mathrm{diag}(\tau^{-1})\cdot x+f(x,u;\theta)$ &
$\times$ &
Euler / ODE solver &
Recurrence / jump updates &
Learned $\tau(x,u)$ \\
CfC \cite{hasani2022closed} &
$x(t{+}\Delta_k){=}(1{-}e^{-\Delta_k/\tau_{\mathrm{eff}}})\odot(f{-}x){+}x$ &
$\times$ &
Closed-form exponential &
Single forward pass &
No ODE solver required \\
NCP \cite{lechner2018neuronal} &
LTC + sparse $W_{\mathrm{rec}}$ (\textit{C.\ elegans}) &
$\times$ &
Euler / ODE solver &
Recurrence &
Connectome-constrained topology \\

Liquid-S4 \cite{hasani2022liquid} &
$\dot{x}=A(u)x+B(u)u$ (liquid $\tau$) &
$\times$ &
ZOH with input-dep.\ params &
Selective scan &
Liquid time-constant in S4 \\

coRNN \cite{rusch2021cornn} &
$\dot{x}=\tanh(\dot{y}+Wx+Bu)$, $\dot{y}=\sigma(W_yx{+}b)-y$ &
$\times$ &
Symplectic Euler &
Recurrence &
Coupled oscillatory dynamics \\
LAN \cite{razzaq2026fluid} &
$\dot{a}=-\mathrm{diag}(\tau^{-1})\cdot a+f(x,u;\theta)$ &
$\times$ &
ODE solver / closed-form &
Attention logit evolution &
ODE-evolved attention logits \\
NAC \cite{razzaq2025neuronal} &
Closed-form approx.\ of LAN-ODE &
$\times$ &
Closed-form approximation &
NCP-wired attention &
\textit{C.\ elegans} attention circuit \\
NSAC \cite{razzaq2026neuronal} &
Gated OU SDE for attention logits &
Gaussian diffusion over attn.\ logits &
Closed-form approximation &
Stochastic attention logits &
Uncertainty estimates in attention \\
\midrule

\multicolumn{6}{c}{\textbf{Family III: Freely Parameterized Vector Fields}}\\
\midrule
NODE \cite{chen2018neural} &
$\dot{x}=\mathrm{MLP}(x,u;\theta)$ &
$\times$ &
Adaptive RK &
Pre-interpolation of $u(t)$ &
Arbitrary neural vector field \\
NCDE \cite{kidger2020neural} &
$dx=f(x;\theta)\,dU(s)$ &
$\times$ &
Adaptive RK over spline $U$ &
Riemann$\times$Stieltjes integral &
Continuous input representation \\
NRDE \cite{morrill2021neuralroughdifferentialequations, lyons2007differential} &
$dx=f(x;\theta)\,d\mathbf{X}(s)$ (rough path) &
$\times$ &
Rough path solver &
Rough path-driven integral &
Handles non-smooth signals \\
NSDE \cite{tzen2019neural,li2020scalable} &
$\dot{x}=\mathrm{MLP}(x,u;\theta)$ &
$\mathrm{MLP}(x,u;\varphi)$ &
Euler$\times$Maruyama / Milstein &
Pre-interpolation &
Learned stochastic diffusion \\

NJSDE \cite{jia2019neural} &
$\mathrm{MLP}(x,u;\theta)$ &
$\mathrm{MLP}(x,u;\varphi)$ + jumps &
Euler with jump process &
Jump-diffusion trajectory &
Jump stochastic dynamics \\

ANODEs \cite{dupont2019augmented, ijcai2019p103} &
$\dot{x}=f([x,\mathrm{aug}];\theta)$ &
$\times$ &
Adaptive RK &
Augmented state output &
Augmented state space \\
SONODEs \cite{bodnar2020second} &
$\ddot{x}=f(x,\dot{x},u;\theta)$ &
$\times$ &
Adaptive RK &
Position and velocity output &
Second-order dynamics \\

LNN \cite{cranmer2020lagrangian} &
Euler$\times$Lagrange from $L(x,\dot{x})$ &
$\times$ &
Symplectic / adaptive RK &
Standard state observation &
Lagrangian structure \\
Graph ODE \cite{poli2021graph} &
$\dot{x}=\mathrm{GNN}(x,\mathcal{A};\theta)$ &
$\times$ &
Adaptive RK &
Node features over time &
Graph-coupled vector field \\
Latent-ODE \cite{rubanova2019latent} &
$\dot{z}=\mathrm{MLP}(z;\theta)$ &
$\times$ &
Adaptive RK &
Encoder$\times$ODE$\times$decoder &
Variational latent initial state \\
ODE-RNN \cite{rubanova2019latent} &
$\dot{x}=\mathrm{MLP}(x;\theta)$ between obs. &
$\times$ &
Adaptive RK + jump &
Hybrid jump updates &
Continuous-discrete hybrid model \\

\midrule

\multicolumn{6}{c}{\textbf{Family IV: Selective State-Space Models}}\\
\midrule
Mamba \cite{gu2023mamba} &
$\dot{x}=Ax+B(u)u$ &
$\times$ &
ZOH with $\Delta_k(u)$ &
Selective scan &
Input-dependent $B,C,\Delta_k$ \\
Mamba-2 \cite{dao2024transformers} &
$\dot{x}=ax+B(u)u$ ($a$ scalar) &
$\times$ &
ZOH with $\Delta_k(u)$ &
SSD scan &
Scalar $A$, attention duality \\

Jamba \cite{lieber2024jamba} &
Mamba + attention + MoE layers &
$\times$ &
ZOH with $\Delta_k(u)$ + attention &
Selective scan + attention &
SSM$\times$attention$\times$MoE hybrid \\
Samba \cite{ren2024samba} &
Mamba + sliding window attention &
$\times$ &
ZOH with $\Delta_k(u)$ &
Selective scan + local attention &
Hybrid SSM-attention mixing \\
Vim \cite{zhu2024vision} &
Bidirectional selective SSM &
$\times$ &
ZOH with $\Delta_k(u)$ &
Bidirectional selective scan &
Vision-adapted bidirectional SSM \\
VMamba \cite{liu2024vmamba} &
2D selective SSM (cross-scan) &
$\times$ &
ZOH with $\Delta_k(u)$ &
2D selective scan (SS2D) &
2D spatial scanning mechanism \\
MambaByte \cite{wang2024mambabyte} &
Selective SSM at byte level &
$\times$ &
ZOH with $\Delta_k(u)$ &
Byte-level selective scan &
No tokenization required \\
MambaML \cite{zhu2025mambaml} &
Selective SSM for general ML &
$\times$ &
ZOH with $\Delta_k(u)$ &
Domain-specific scan &
Beyond language and vision \\
RetNet \cite{sun2023retentive} &
$\dot{x}=\mathrm{diag}(\gamma)x+u$ &
$\times$ &
Parallel / recurrence &
Retention recurrence &
Fixed learnable decays $\gamma$ \\
RWKV \cite{peng2023rwkv} &
Input-dependent linear recurrence &
$\times$ &
Linear recurrence (parallel) &
Token-shifted recurrence &
Linear RNN with time-mixing \\
Griffin \cite{de2024griffin} &
Input-dep.\ gated linear recurrence &
$\times$ &
Parallel scan &
Recurrence + local attention &
Hybrid gated recurrence \\
GLA \cite{yang2023gated} &
Gated linear attention dynamics &
$\times$ &
Chunkwise parallel / recurrence &
Linear attention &
Input-dependent gating in attention \\
S7 \cite{soydan2024s7selectivesimplifiedstate} &
$\dot{x}=Ax+Bu+\sigma(x)$ &
$\times$ &
ZOH &
Parallel scan &
Elementwise nonlinearity in SSM \\

\midrule
\multicolumn{6}{c}{\textbf{Family V: Continuous-Time Attention Models}}\\
\midrule
mTAN \cite{shukla2021multi} &
trajectory-free member &
$\times$ &
Direct weighted sum &
$\sum_k\alpha(t_q,t_k)u_k$ &
Learned continuous-time kernel \\
ANCDE \cite{jhin2021ace,jhin2024attentive} &
$\frac{d\alpha}{dt}=f_\alpha(\alpha;\theta_\alpha)\,dU$, $\frac{dx}{dt}=f_x(x,\alpha;\theta_x)\,dU$ &
$\times$ &
Adaptive RK over spline $U$ &
Dual NCDE coupling &
Attention as co-evolving dynamical system \\
ContiFormer \cite{chen2023contiformer} &
$\dot{z}_i=f(z_i;\theta)$ (per element) &
$\times$ &
Adaptive RK + attention &
Continuous $Q,K,V$ trajectories &
Trajectory-evolved attention \\
OT-Transformer \cite{kan2025ot} &
ODE for full transformer block &
$\times$ &
Adaptive RK &
ODE-evolved attention &
OT-regularized ODE transformer \\
PDE-Attention \cite{zhang2025continuous} &
PDE over attention matrix (pseudo-$t$) &
$\times$ &
PDE solver &
PDE-evolved attention matrix &
PDE over attention weights \\
CTA~\cite{chien2021continuous} &
Continuous relaxation of softmax &
$\times$ &
Integral approximation &
Continuous weighted kernel integral &
Continuous attention distribution \\
\bottomrule
\end{tabular}%
}
\\
\begin{minipage}{1.4\textwidth}
\footnotesize
\vspace{1mm}
\textbf{Abbreviations:}
RK = Runge$\times$Kutta; EMA = Exponential Moving Average; FFT = Fast Fourier Transform; 
GLA = Gated Linear Attention; ANODE = Augmented ODE; LSSL = Linear State-Space Layer; LRU = Linear Recurrent Unit; GSS = Gated State Space; BiGS = Bidirectional Gated SSM;  MoE = Mixture-of-Experts
\end{minipage}
\end{table}
\end{landscape}

\subsubsection{Backpropagation through time (BPTT)}
BPTT~\cite{rumelhart1986learning} unrolls the discretized recurrent dynamics over $n$ time steps into a feedforward computation graph, then computes gradients through that graph. For a dense recurrent model with $k$ hidden units, BPTT costs $\mathcal{O}(nk^2)$ in computation and $\mathcal{O}(nk)$ in memory, because every intermediate hidden state must be saved for the backward pass. Irregularly sampled observations need varying time intervals to be incorporated into the dynamics, either through continuous time integration, adaptive solver steps, or a timestamp-aware formulation. Alternatively, resampling onto a uniform grid simplifies computation but may introduce interpolation errors. The primary limitation of BPTT is its linear memory growth with sequence length, which becomes challenging for very long temporal sequences.

\subsubsection{Adjoint Sensitivity Methods}

Adjoint sensitivity methods reduce memory requirements by computing gradients through an additional backward-in-time adjoint differential equation:
\begin{equation}
\dot{a}(t)=-\left(\frac{\partial f}{\partial x}\right)^{\mathsf{T}}a(t),\qquad a(T)=\frac{\partial L}{\partial x(T)},\qquad \frac{\partial L}{\partial\theta}=\int_{t_0}^{T}a(t)^{\mathsf{T}}\frac{\partial f}{\partial\theta}\,dt
\end{equation}
It assumes a terminal loss $L = \ell(x(T))$ that depends only on the terminal state. For a running loss $L = \int_{t_0}^{T} \ell(x(t),t),dt + \ell(x(T))$, the adjoint equation includes an additional source term $-\partial \ell/\partial x$ on the right-hand side. The memory cost drops to roughly $\mathcal{O}(1)$ in the number of solver steps, but it introduces additional numerical computational cost that may increase training time compared with checkpointing-based approaches. Gholaminejad et al.~\cite{ijcai2019p103} showed that on typical NODE training workloads, the adjoint method costs about as much as truncated BPTT with checkpointing, or more; checkpointing recomputes a few forward states instead of storing every one. The ANODE framework~\cite{dupont2019augmented}  improves gradient reliability by addressing numerical reconstruction error during backward integration through augmented dynamics and reversible integration strategies. In most practical settings, discretize-then-optimize plus checkpointing provides a favorable point on the memory-accuracy curve.

\subsubsection{Parallel Associative Scan}
The parallel associative scan algorithm uses the associative structure of linear state-space recurrence to perform efficient parallel computation.  A state update $x_k = \bar{A}_k x_{k-1} + \bar{B}_k u_k$ is an affine pair $(p,q)$ with composition $(a_2, b_2) \circ (a_1, b_1) = (a_2 a_1, a_2 b_1 + b_2)$. The operation is associative, so all $n$ prefix transformations can be computed in parallel with $\mathcal{O}(n)$ work and $\mathcal{O}(\log n)$ depth on $\mathcal{O}(n)$ processors. For S4, specialized parameterizations of the transition matrix reduce the computational complexity compared with dense matrix operations. Mamba adds input-dependent (selective) state-space parameters and keeps the associative scan because each transition is still affine. On GPU architectures, these operations are implemented using optimized CUDA kernels that exploit parallelism across sequences and channel dimensions.

\subsection{The Vanishing Gradient Problem}
The vanishing gradient problem is a fundamental limitation of all recurrent architectures. CT models have different approaches for dealing with it than DT models. In \emph{Family I \& IV}, the HiPPO (specifically LegS) matrix is a structured real matrix whose eigenvalues lie on the negative real axis; under diagonal rescaling it is similar to a normal matrix, and under bilinear or ZOH discretization the transition eigenvalues lie in $(0,1)$, bounded away from 0 for small $\Delta_k$. The result is slow, non-vanishing decay that holds long-range memory without learned gating; bilinear discretization preserves CT stability by mapping the open left half-plane into the interior of the unit circle. Euler discretization fails to preserve this norm-preservation property, so S4 and Mamba use ZOH or bilinear discretization instead. \emph{Family II} addresses gradient propagation through the input-dependent time constant $\tau(x,u)$ adaptively controlling the rate of state evolution, and allowing dimensions requiring long-term memory to evolve slowly. 
This mechanism is analogous to the LSTM forget gate, but achieves memory retention through CT dynamics rather than an explicit multiplicative gate. CfC and NAC's closed form solutions stabilize training by bounding the state update through the exponential decay ($1 - e^{-\Delta_k/\tau_{\text{eff}}}$), which prevents unbounded state growth. 
For \emph{Family III}, stiffness is the continuous analogous of vanishing gradients. When the eigenvalues of $\partial f/\partial x$ span orders of magnitude, the ODE solver is forced to take impractically small steps, and the adjoint gradient becomes inaccurate. Finlay et al.~\cite{finlay2020train} regularize the Jacobian norm during training to address stiffness. Kelly et al.~\cite{kelly2020learning} reparameterize the dynamics to satisfy Lipschitz constraints. 

\pagebreak

\begin{wrapfigure}{r}{0.5\columnwidth}
\centering
\vspace{3mm}
\centering
\includegraphics[width=0.49\textwidth]{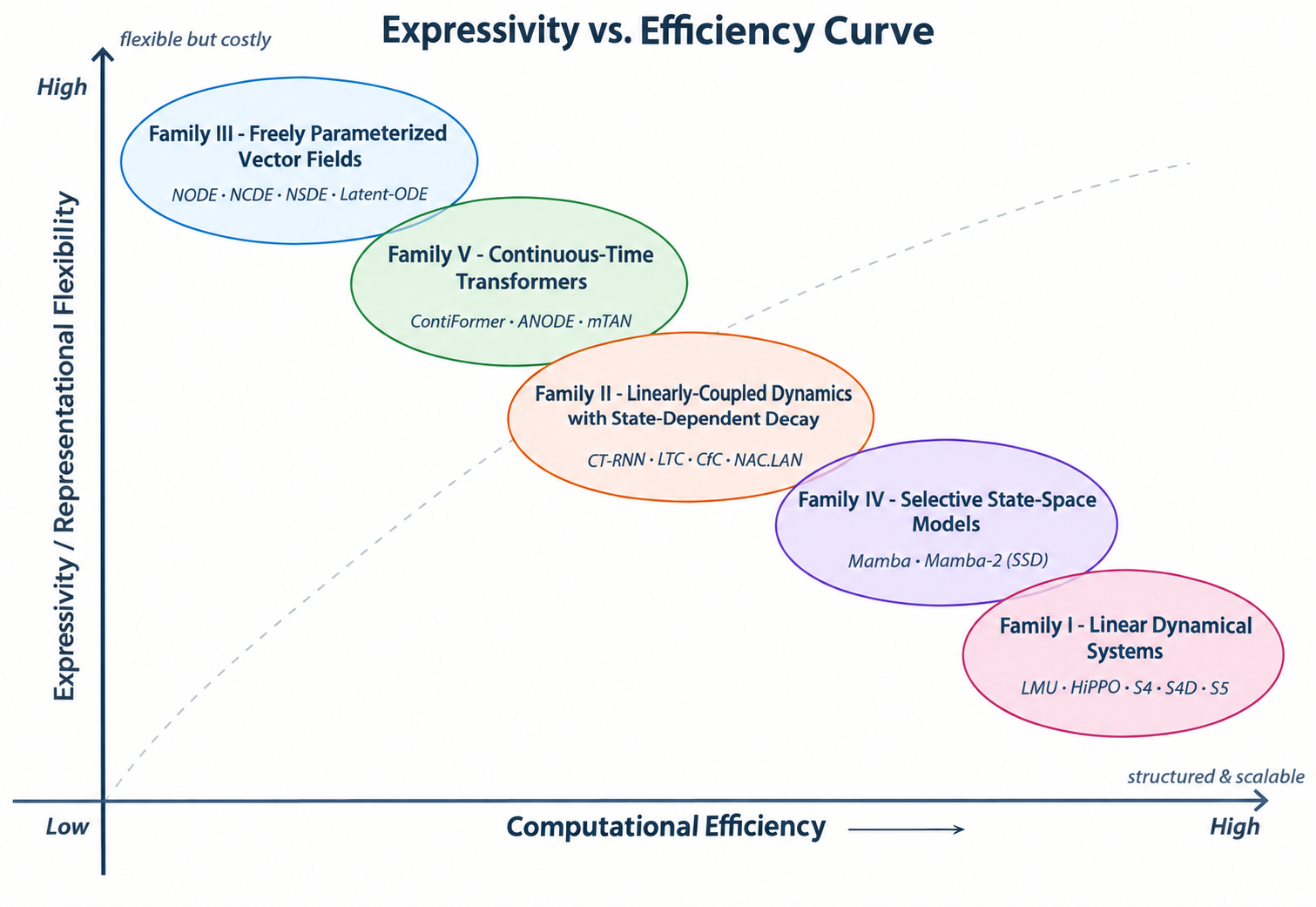}
\caption{Conceptual expressivity-efficiency trade-off landscape across families.}
\label{fig:expressivity}
\vspace{-8mm}
\end{wrapfigure}

\subsection{Failure Modes and Trade-offs}

\textit{\textbf{Expressivity vs. Computational Complexity:}} \emph{Family III} provides the highest representational flexibility; it parameterizes a general smooth vector field with few structural restrictions. However, this flexibility comes with a significant computational overhead, as numerical integration requires $S$ evaluations of the learned dynamics per integration step, with the number of evaluations potentially ranging from tens to thousands in stiff regimes. \emph{Family I \& IV} impose linear dynamical assumptions, resulting in reduced expressivity but favorable computational complexity. \emph{Family II} occupies an intermediate position by restricting the vector field structure through leaky-integrator dynamics and closed-form approximations, achieving improved efficiency while retaining nonlinear temporal modeling capacity. \emph{Family V} adds adaptive interactions on top of these dynamics but pays the quadratic attention cost in sequence length. 
Figure~\ref{fig:expressivity} illustrates the expressivity vs. computational efficiency curve.

\textbf{\textit{Stiffness \& Solver-Induced Computational Failure:}} A fundamental limitation of \emph{Family III} training is the solver-induced stiffness. During optimization, the learned vector field may converge to regions where the Jacobian spectrum $\partial f/\partial x$ exhibits widely separated eigenvalues. This is a computational failure, not a representational one: the dynamics can still represent the data, but solving them numerically becomes too expensive. \emph{Family I \& IV} mitigate this issue by explicitly controlling system dynamics through structured state transitions. \emph{Family II} avoids adaptive integration instability through its closed-form update formulation (i.e., CfC and NAC).

\subsection*{Key Takeaways}
Training approaches are divided into three algorithms with distinct computational and memory trade-offs. BPTT stores all intermediate states, making it computationally expensive for long sequences. The adjoint method reduces memory usage to a constant order but requires solving a backward differential equation, whose accuracy can be sensitive to numerical errors; moreover, its memory advantages over discrete checkpointing diminish when advanced checkpointing strategies are employed. Parallel associative scan trades the sequential dependency for linear-time computation, enabling the scalable training of SSMs. Stiff vector fields are analogous to vanishing gradients in their discrete counterparts. Stiffness is characterized by eigenvalues spanning several orders of magnitude and can force adaptive solvers to take impractically small step sizes. The key takeaway is that most practical limits of CT models come from computation, not representation. An unconstrained vector field can in principle represent the desired dynamics, but the solver cost makes it unusable. That is why structured and closed-form alternatives win in practice.

\begin{table}[t]
\centering
\caption{Per-layer complexity; \textit{Training costs}: per sequence of length $n$. \textit{Inference costs}: per generated step. $n$: sequence length; $k$: hidden/state dimension; $d$: attention width ($d_{\text{model}}{=}hd$, heads absorbed); $S$: ODE-solver evaluations steps ($S{=}1$) for solver-free/closed-form models); $w$: attention window.}
\vspace{1mm}
\label{tab:model_complexity}
\small
\resizebox{0.9\textwidth}{!}{%
\begin{tabular}{m{5.6cm}ccccc}
\toprule
\textbf{Model} &
\textbf{Train time} & \textbf{Train memory} &
\textbf{Infer time} & \textbf{Infer memory} & \textbf{Parallel} \\
& \textit{(per seq. $n$)} & \textit{(per seq. $n$)} &
\textit{(per step)} & \textit{(per step)} & \textbf{training} \\
\midrule
\multicolumn{6}{c}{\textbf{Family I: Linear Dynamical Systems}}\\
\midrule
LMU, LSSL, S4, S4D/DSS, GSS, H3, BiGS, S4ND, SaShiMi
& $\mathcal{O}(nk\log n)$ & $\mathcal{O}(nk)$ & $\mathcal{O}(k)$ & $\mathcal{O}(k)$ & Yes \\
LRU, S5
& $\mathcal{O}(nk)$ & $\mathcal{O}(nk)$ & $\mathcal{O}(k)$ & $\mathcal{O}(k)$ & Yes \\
\midrule
\multicolumn{6}{c}{\textbf{Family II: Linearly-Coupled Dynamics with State-Dependent Decay}}\\
\midrule
LSTM, PhasedLSTM, coRNN, CfC
& $\mathcal{O}(nk^2)$ & $\mathcal{O}(nk)$ & $\mathcal{O}(k^2)$ & $\mathcal{O}(k)$ & No \\
CT-RNN, LTC, Liquid-S4
& $\mathcal{O}(nSk^2)$ & $\mathcal{O}(k)^{\dag}$ & $\mathcal{O}(Sk^2)$ & $\mathcal{O}(k)$ & No \\
NCP
& $\mathcal{O}(nSk)$ & $\mathcal{O}(k)^{\dag}$ & $\mathcal{O}(Sk)$ & $\mathcal{O}(k)$ & No \\
CT-GRU
& $\mathcal{O}(nk^2)$ & $\mathcal{O}(nk)$ & $\mathcal{O}(k^2)$ & $\mathcal{O}(k)$ & No \\
LAN
& $\mathcal{O}(n^2Sd)$ & $\mathcal{O}(nd)$ & $\mathcal{O}(n^2)$ & $\mathcal{O}(nd)$ & Yes \\
NAC
& $\mathcal{O}(n\sqrt{n}\,d)$ & $\mathcal{O}(n\sqrt{n})$ & $\mathcal{O}(nd)$ & $\mathcal{O}(nd)$ & Yes \\
NSAC
& $\mathcal{O}(nS\sqrt{n}\,d)$ & $\mathcal{O}(nSd)$ & $\mathcal{O}(nS\sqrt{n})$ & $\mathcal{O}(nd)$ & Yes \\
\midrule
\multicolumn{6}{c}{\textbf{Family III: Freely Parameterized Vector Fields}}\\
\midrule
NODE/ANODE/SONODE/HNN/LNN
& $\mathcal{O}(nSk^2)$ & $\mathcal{O}(k)^{\dag}$ & $\mathcal{O}(Sk^2)$ & $\mathcal{O}(k)$ & No \\
NCDE/NRDE
& $\mathcal{O}(nSk^2)$ & $\mathcal{O}(nd_{\text{in}})$ & $\mathcal{O}(Sk^2)$ & $\mathcal{O}(nd_{\text{in}})$ & No \\
NSDE / JSDE
& $\mathcal{O}(nSk^2)$ & $\mathcal{O}(k)^{\dag}$ & $\mathcal{O}(Sk^2)$ & $\mathcal{O}(k)$ & No \\
Graph ODE
& $\mathcal{O}(S|E|k)$ & $\mathcal{O}(|V|k)^{\dag}$ & $\mathcal{O}(S|E|k)$ & $\mathcal{O}(|V|k)$ & No \\
Latent ODE
& $\mathcal{O}(nSk^2)$ & $\mathcal{O}(nk)$ & $\mathcal{O}(Sk^2)$ & $\mathcal{O}(k)$ & No \\
ODE-RNN
& $\mathcal{O}(nSk^2)$ & $\mathcal{O}(nk)^{\dag}$ & $\mathcal{O}(Sk^2)$ & $\mathcal{O}(k)$ & No \\
\midrule
\multicolumn{6}{c}{\textbf{Family IV: Selective State-Space Models}}\\
\midrule
Mamba, Vim/VMamba, MambaByte / MambaML, S7, Mamba-2, RetNet, RWKV, GLA
& $\mathcal{O}(nk)$ & $\mathcal{O}(nk)$ & $\mathcal{O}(k)$ & $\mathcal{O}(k)$ & Yes \\
Jamba 
& $\mathcal{O}(n^2 d)$ & $\mathcal{O}(nk)$ & $\mathcal{O}(nd)$ & $\mathcal{O}(nd)$ & Yes \\
Samba, Griffin
& $\mathcal{O}(nwd)$ & $\mathcal{O}(nk)$ & $\mathcal{O}(wd)$ & $\mathcal{O}(wd)$ & Yes \\
\midrule
\multicolumn{6}{c}{\textbf{Family V: Continuous-Time Transformers}}\\
\midrule
mTAN/CTA
& $\mathcal{O}(n^2 d)$ & $\mathcal{O}(n^2)$ & $\mathcal{O}(nd)$ & $\mathcal{O}(nd)$ & Yes \\
ANCDE
& $\mathcal{O}(nSk^2)$ & $\mathcal{O}(nd_{\text{in}})$ & $\mathcal{O}(Sk^2)$ & $\mathcal{O}(nd_{\text{in}})$ & Yes \\
ContiFormer
& $\mathcal{O}(n^2 d)$ & $\mathcal{O}(n^2)$ & $\mathcal{O}(nd)$ & $\mathcal{O}(nd)$ & Yes \\
OT-Transformer, PDE-Attention
& $\mathcal{O}(Sn^2 d)$ & $\mathcal{O}(n^2)$ & $\mathcal{O}(Snd)$ & $\mathcal{O}(nd)$ & Yes \\
\bottomrule
\end{tabular}}
\begin{minipage}{0.9\textwidth}
\footnotesize
\textbf{Note:} \dag~training memory assumes the adjoint method (unfolded BPTT costs $\mathcal{O}(nSk)$). Absorbed as constants: SSM state order, scan/chunk size, DPLR rank $r$; Graph edges $|E|$ and node $|V|$; NCP sparsity $s{\approx}15$, Hutchinson probes. Pointwise projections/FFN ($\mathcal{O}(k^2)$/token), common to all families, are excluded.
\end{minipage}
\end{table}

\section{Comparative Analysis \& Ecosystems}\label{section:benchmarks}

\subsection{Comparative Analysis}
In this section, we first provide a theoretical computational complexity analysis; then we perform an illustrative architecture-controlled benchmark analysis on the representative architectures from each family.

\subsubsection{Complexity Analysis}
Table~\ref{tab:model_complexity} provides a comprehensive comparison of the computational and memory complexities across all families. A recurring pattern across sequence modeling architectures is the trade-off between training parallelism and inference efficiency. \emph{Family V} provides full parallelism across sequence positions during training; however, self-attention introduces quadratic computational and memory complexity with respect to sequence length, resulting in $\mathcal{O}(n^2d)$ computation and $\mathcal{O}(n^2)$ memory for attention operations.\\
\emph{Family I \& IV} achieve near-linear training complexity by reformulating recurrent state updates as parallel associative scans. During autoregressive inference, these models maintain only a fixed-dimensional recurrent state, requiring $\mathcal{O}(k)$ memory. This efficiency is obtained by sacrificing some training parallelism, as scan-based operations introduce synchronization dependencies compared with fully parallel attention mechanisms. Specifically, \emph{Family IV} achieves near-linear training cost by restricting the parallelism overhead to the scan operation while preserving efficient recurrent inference.\\
The adjoint-trained \emph{Family II \& III} achieve memory-efficient state evolution without requiring the storage of all intermediate activations. However, their inference process relies on sequential numerical integration of the underlying dynamics, which can require a potentially large number of function evaluations, $S$, particularly for stiff systems. Despite this sequential solver cost, NDE-based models achieve favorable asymptotic inference memory complexity of $\mathcal{O}(k)$, since the recurrent formulation propagates hidden states without additional computation at time points where no observations are available; their per-step inference time remains $\mathcal{O}(Sk^2)$. 

\subsubsection{Benchmark Analysis}\label{section:empirical_analysis}
The empirical benchmarking of CT machine learning is fragmented, with different research communities adopting different datasets, evaluation protocols, preprocessing pipelines, and computational budgets to assess their proposed architectures. As a result, it is often difficult to directly compare different model families on the same benchmark dataset based solely on results reported in the original papers. Establishing a comprehensive unified benchmark that evaluates all representative architectures under the same constraints remains an open challenge. 

\begin{wrapfigure}{r}{0.55\textwidth}
\captionsetup{type=table}
% \vspace{-10mm}
\centering
\caption{Architecture-Specific Solver/Discretization Settings.}
\label{tab:model_hparams}
\resizebox{0.53\textwidth}{!}{%
\begin{tabular}{@{}llp{6.5cm}@{}}
\toprule
\textbf{Family} & \textbf{Model} & \textbf{Solver / Discretization} \\
\midrule
\multirow{4}{*}{\textbf{I}}
& DeepState & Linear recurrence \\
& S4        & ZOH + FFT convolution \\
& HiPPO     & ZOH / linear recurrence (HiPPO--LegS) \\
& S5        & ZOH + recurrent scan (multi-head diagonal) \\
\midrule
\multirow{6}{*}{\textbf{II}}
& CT-RNN & Euler  \\
& CT-GRU & Closed-form \\
& LTC    & Fused Solver (default) \\
& CfC    & Closed-form \\
& FLUID  & Euler (default) \\
& NAC    & Closed-form \\
\midrule
\multirow{5}{*}{\textbf{III}}
& NODE/ODE-LSTM  & Euler  \\
& NCDE  & Euler--Riemann--Stieltjes \\
& NRDE  & Rough-path controlled ODE \\
& NSDE  & Euler--Maruyama  \\
\midrule
\multirow{4}{*}{\textbf{IV}}
& Mamba  & ZOH + parallel associative scan \\
& S7     & ZOH + scan (elementwise nonlinearity) \\
& Jamba  & ZOH + scan; Mamba + attention + MoE \\
& RetNet & Retention recurrence / parallel view \\
\midrule
\multirow{3}{*}{\textbf{V}}
& ContiFormer / ODEFormer & Euler \\
& OT-Transformer & Euler \\
& PDE-Attention & PDE evolution \\
& mTAN & $\times$ \\
\bottomrule
\end{tabular}}
\end{wrapfigure}

While creating such a comprehensive benchmark is beyond the scope of this survey, we perform an illustrative evaluation of representative architectures under a shared restricted standardized configuration to complement the conceptual analysis presented throughout the survey. The objective is to compare architectures under a common optimization and architecture-controlled configuration: layer count, hidden width, feed-forward width, optimizer, schedule, and seed are held fixed. Total parameters counts are not equalized because attention projections, SSM state dimensions, and ODE-field MLPs scale differently; capacity therefore remains a confound, and the results should be read as illustrating relative trade-off among algorithmic formulation under on restricted setting rather than isolating formulation from scalability.\\
We considered three experiments: (i) irregular time-series modeling; (ii) long-range forecasting; and (iii) runtime analysis. For the irregular time-series experiments, we selected two benchmark datasets: (a) PhysioNet~\cite{physionet} and (b) Event-based MNIST (E-MNIST)~\cite{lechner2022mixed}. For long-range forecasting, we used the ETTm1 and Jena Climate datasets. 

\textbf{\textit{Preprocessing:}} To transform the MNIST~\cite{lecun1998gradient} dataset into an event-based dataset, we followed the three-step preprocessing procedure described in~\cite{lechner2022mixed}. \textit{First}, a threshold is applied to binarize the 8-bit pixel intensities (range 0--255). \textit{Second}, each $28 \times 28$ image is reshaped into a 1-D time-series of length 784. The resulting binary sequence is then encoded into an event-based representation by removing consecutive occurrences of the same value (e.g., 1,1,1,1 $\to$ 1, t=4). This introduces a temporal structure and also compresses the sequence length from 784 to an average of 53 time steps. \textit{Third}, each sequence is padded to a fixed length of 256. The time dimensions are normalized such that each event corresponds to one unit of time. For PhysioNet, we followed the procedure described in~\cite{johnson2012patient}. ETTm1 and Jena Climate are long-range multivariate time-series datasets used for forecasting. Prior to training, we normalized the input features of both datasets using \texttt{StandardScaler}.

\textbf{\textit{Competing architectures:}} We evaluated prominent representative architectures from each family: \emph{Family I} $\in \{\text{HiPPO, DeepState, S4, S5}\}$; \emph{Family II} $\in \{\text{CT-RNN, CT-GRU, PhasedLSTM, LTC, CfC, FLUID, NAC}\}$; \emph{Family III} $\in \{\text{NODE, NCDE, NRDE, NSDE, ODE-LSTM}\}$; \emph{Family IV} $\in \{\text{Mamba, Jamba, S7, RetNet}\}$; and \emph{Family V} $\in \{\text{mTAN, PDE-Attention, ODEFormer, ContiFormer, OT-Transformer}\}$. We also considered DT-models $\in \{\text{LSTM, GRU, SDPA-Transformer}\}$ in our analysis. All experiments were run in TensorFlow; for models whose reference kernels are not available, we re-implemented them to the best of our ability, following the formulations of Table~\ref{tab:ct_taxonomy}. Small deviations from the official kernels are therefore expected.

\textbf{\textit{Experimental Setup:}}  To compare architectures under a common experimental protocol, we used the same optimization and architecture-controlled hyperparameters across all experiments, rather than independently tuning these settings for each model. Specifically, all models were trained with a constant learning rate of $0.001$, zero weight decay, a batch size of 32, a maximum of 50 training epochs, no learning-rate scheduler, and no early stopping, with the random seed fixed to 42. The architecture-controlled settings were also held fixed across experiments, including a single model layer ($n_{\mathrm{layers}}=1$), hidden dimension $d_n=64$, feed-forward dimension $d_f=64$, and, where applicable, $d_h=16$ attention heads, $d_{\mathrm{state}}=16$ for S7 and Jamba, with HiPPO and S5 using the full hidden dimension as the state dimension, $d_{\mathrm{conv}}=4$ for S7 and Jamba, $\mathrm{topk}=8$ for FLUID and NAC, PDE parameters $\epsilon_\alpha=0.1$ and $\tau_{pde}=0.1$. Solver and discretization choices were selected with computational efficiency as the priority and then fixed across experiments; these settings are summarized in Table~\ref{tab:model_hparams}. This protocol ensures that differences in performance are evaluated under a common optimization and architecture configuration, while allowing solver and discretization choices to reflect computational-efficiency considerations.\\
For each experiment, we first performed an 80:20 chronological split into a training set and held-out test set. The training portion was then partitioned into five folds, with four folds used for training and one for validation in each run. Training was performed end-to-end using BPTT. Each fold-specific model was evaluated on the same held-out test set, and we report the mean and standard deviation across the five fold-specific evaluations. PhysioNet performance was quantified using the area under the receiver operating characteristic curve (AUC), E-MNIST performance using classification accuracy, and long-range forecasting performance using mean squared error (MSE) at the specified forecasting horizons. All experiments were performed on a system with an Intel Core i9-9900X CPU, 64~GB RAM, and a single NVIDIA RTX 3090 GPU. The software environment used TensorFlow 2.18.0, with GPU memory growth enabled.

\begin{table}[t]
\centering
\caption{Empirical and computational results.}
\label{tab:all_results}
\vspace{1mm}
\resizebox{\textwidth}{!}{%
\begin{tabular}{lccccccccccc}
\toprule

\multirow{2}{*}{\textbf{Family}} &
\multirow{2}{*}{\textbf{Model}} &
\multicolumn{2}{c}{\textbf{Irregular Time-Series}} &
\multicolumn{2}{c}{\textbf{Long-Range Forecasting}$^{\dagger}$} &
\multicolumn{5}{c}{\textbf{Runtime \& Memory}$^{\ddagger}$} \\

\cmidrule(lr){3-4}
\cmidrule(lr){5-6}
\cmidrule(lr){7-11}
&
& \textbf{E-MNIST~($\uparrow$)}
& \textbf{PhysioNet~($\uparrow$)}
& \textbf{ETTm1~($\downarrow$)}
& \textbf{Jena-Climate~($\downarrow$)}
& \makecell{\textbf{RunTime}~($\downarrow$)\\(s)} 
& \makecell{\textbf{Throughput}~($\uparrow$)\\(seq/s)} 
& \makecell{\textbf{PeakMem}~($\downarrow$)\\(MB)}
& \makecell{\textbf{FLOPS}~($\downarrow$)\\($\times10^{9}$)}
& \makecell{\textbf{Params.}~($\downarrow$)\\($\times10^{3}$)}\\

\midrule

\multirow{3}{*}{\textbf{DT}}
& LSTM
& 89.19\textsuperscript{\scriptsize $\pm$2.90}
& 0.5218\textsuperscript{\scriptsize $\pm$0.1066}
& 1.2507\textsuperscript{\scriptsize $\pm$0.3610}
& 2.3611\textsuperscript{\scriptsize $\pm$0.6838}
& 2.9648\textsuperscript{\scriptsize $\pm$0.4824}
& 0.34
& 0.43 
& 0.068 
& 33.0 \\

& GRU
& 91.56\textsuperscript{\scriptsize $\pm$0.92}
& 0.7993\textsuperscript{\scriptsize $\pm$0.0206}
& 0.1336\textsuperscript{\scriptsize $\pm$0.0188}
& 1.2670\textsuperscript{\scriptsize $\pm$0.1250}
& 3.8572\textsuperscript{\scriptsize $\pm$0.4648}
& 0.26
& 0.53 
& 0.051 
& 25.0 \\

& SDPA-Transformer
& 93.70\textsuperscript{\scriptsize $\pm$0.19}
& 0.6564\textsuperscript{\scriptsize $\pm$0.0313}
& 0.1918\textsuperscript{\scriptsize $\pm$0.0185}
& 3.1536\textsuperscript{\scriptsize $\pm$0.5416}
& 0.0122\textsuperscript{\scriptsize $\pm$0.0014}
& 82.25
& 321.63 
& 0.362 
& 29.4 \\

\midrule

\multirow{4}{*}{\textbf{I}}
& HiPPO
& 94.21\textsuperscript{\scriptsize $\pm$0.47}
& 0.6985\textsuperscript{\scriptsize $\pm$0.0225}
& 0.0544\textsuperscript{\scriptsize $\pm$0.0025}
& 1.6204\textsuperscript{\scriptsize $\pm$0.3437}
& 0.3854\textsuperscript{\scriptsize $\pm$0.0373}
& 2.59
& 2.21 
& 0.017 
& 8.4 \\

& DeepState
& 94.77\textsuperscript{\scriptsize $\pm$0.25}
& 0.513\textsuperscript{\scriptsize $\pm$0.0123}
& 0.1182\textsuperscript{\scriptsize $\pm$0.0105}
& 1.1887\textsuperscript{\scriptsize $\pm$0.1501}
& 0.4867\textsuperscript{\scriptsize $\pm$0.0742}
& 2.05
& 2.00
& 0.009 
& 8.3 \\

& S4
& 95.36\textsuperscript{\scriptsize $\pm$0.20}
& 0.6258\textsuperscript{\scriptsize $\pm$0.0279}
& 0.0664\textsuperscript{\scriptsize $\pm$0.0037}
& 2.4713\textsuperscript{\scriptsize $\pm$0.9557}
& 0.0215\textsuperscript{\scriptsize $\pm$0.0082}
& 46.44
& 66.75 
& 0.068 
& 24.7 \\

& S5
& 95.23\textsuperscript{\scriptsize $\pm$0.45}
& 0.5809\textsuperscript{\scriptsize $\pm$0.0308}
& 0.0813\textsuperscript{\scriptsize $\pm$0.0175}
& 2.6426\textsuperscript{\scriptsize $\pm$0.3023}
& 0.5874\textsuperscript{\scriptsize $\pm$0.0420}
& 1.70
& 162.16 
& 0.013 
& 10.4 \\

\midrule

\multirow{7}{*}{\textbf{II}}
& CT-RNN
& 97.51\textsuperscript{\scriptsize $\pm$1.06}
& 0.6708\textsuperscript{\scriptsize $\pm$0.0335}
& 0.1072\textsuperscript{\scriptsize $\pm$0.0130}
& 0.9270\textsuperscript{\scriptsize $\pm$0.1756}
& 12.0392\textsuperscript{\scriptsize $\pm$0.3364}
& 0.08
& 0.54 
& 0.204 
& 24.8 \\

& CT-GRU
& 98.56\textsuperscript{\scriptsize $\pm$0.95}
& 0.8368\textsuperscript{\scriptsize $\pm$0.0349}
& 0.1330\textsuperscript{\scriptsize $\pm$0.0248}
& 1.4513\textsuperscript{\scriptsize $\pm$0.1842}
& 6.4293\textsuperscript{\scriptsize $\pm$0.7540}
& 0.16
& 0.30 
& 0.086 
& 8.3 \\

& PhasedLSTM
& 98.72\textsuperscript{\scriptsize $\pm$1.04}
& 0.8252\textsuperscript{\scriptsize $\pm$0.0113}
& 0.1488\textsuperscript{\scriptsize $\pm$0.0217}
& 1.3396\textsuperscript{\scriptsize $\pm$0.1711}
& 5.2351\textsuperscript{\scriptsize $\pm$0.9880}
& 0.19
& 0.45 
& 0.068 
& 33.0 \\

& LTC
& 93.00\textsuperscript{\scriptsize $\pm$0.70}
& 0.5552\textsuperscript{\scriptsize $\pm$0.0676}
& 0.1096\textsuperscript{\scriptsize $\pm$0.0059}
& 1.6448\textsuperscript{\scriptsize $\pm$0.4434}
& 12.7050\textsuperscript{\scriptsize $\pm$0.8152}
& 0.08
& 0.76 
& 0.073 
& 33.1 \\

& CfC
& 96.10\textsuperscript{\scriptsize $\pm$1.04}
& 0.6552\textsuperscript{\scriptsize $\pm$0.0578}
& 0.0873\textsuperscript{\scriptsize $\pm$0.0064}
& 2.5352\textsuperscript{\scriptsize $\pm$0.9702}
& 15.3939\textsuperscript{\scriptsize $\pm$0.4380}
& 0.06
& 0.41 
& 0.034 
& 22.2 \\

& FLUID
& 95.82\textsuperscript{\scriptsize $\pm$0.23}
& 0.7991\textsuperscript{\scriptsize $\pm$0.0162}
& 0.2305\textsuperscript{\scriptsize $\pm$0.0342}
& 2.8615\textsuperscript{\scriptsize $\pm$0.1002}
& 10.99\textsuperscript{\scriptsize $\pm$0.79}
& 0.09
& 139.71 
& 2.30 
& 182.8 \\

& NAC
& 98.02\textsuperscript{\scriptsize $\pm$0.58}
& 0.7229\textsuperscript{\scriptsize $\pm$0.0165}
& 0.0799\textsuperscript{\scriptsize $\pm$0.0083}
& 1.8055\textsuperscript{\scriptsize $\pm$0.4259}
& 10.4131\textsuperscript{\scriptsize $\pm$1.9616}
& 0.10
& 288.25
& 0.867 
& 177.3 \\

\midrule

\multirow{5}{*}{\textbf{III}}
& NODE
& 97.54\textsuperscript{\scriptsize $\pm$0.51}
& 0.7296\textsuperscript{\scriptsize $\pm$0.0247}
& 0.1197\textsuperscript{\scriptsize $\pm$0.0880}
& 1.6437\textsuperscript{\scriptsize $\pm$0.5251}
& 9.8369\textsuperscript{\scriptsize $\pm$1.0491}
& 0.10
& 0.40 
& 0.127
& 12.4 \\

& NCDE
& 94.95\textsuperscript{\scriptsize $\pm$0.19}
& 0.5561\textsuperscript{\scriptsize $\pm$0.0710}
& 0.1128\textsuperscript{\scriptsize $\pm$0.0055}
& 2.5883\textsuperscript{\scriptsize $\pm$0.3698}
& 6.9534\textsuperscript{\scriptsize $\pm$0.6464}
& 0.14
& 1.49 
& 0.127 
& 12.4 \\

& NRDE
& 94.96\textsuperscript{\scriptsize $\pm$0.12}
& 0.5158\textsuperscript{\scriptsize $\pm$0.0324}
& 0.1184\textsuperscript{\scriptsize $\pm$0.0111}
& 2.7151\textsuperscript{\scriptsize $\pm$0.7411}
& 6.6240\textsuperscript{\scriptsize $\pm$0.1905}
& 0.15
& 2.03 
& 0.128 
& 12.4 \\

& NSDE
& 95.32\textsuperscript{\scriptsize $\pm$0.13}
& 0.7458\textsuperscript{\scriptsize $\pm$0.0236}
& 0.1252\textsuperscript{\scriptsize $\pm$0.0170}
& 1.7885\textsuperscript{\scriptsize $\pm$0.4320}
& 13.7565\textsuperscript{\scriptsize $\pm$0.5207}
& 0.07
& 1.12 
& 0.256 
& 24.8 \\

& ODE-LSTM
& 95.30\textsuperscript{\scriptsize $\pm$0.10}
& 0.7992\textsuperscript{\scriptsize $\pm$0.0279}
& 0.1453\textsuperscript{\scriptsize $\pm$0.0099}
& 1.5855\textsuperscript{\scriptsize $\pm$0.1427}
& 6.0792\textsuperscript{\scriptsize $\pm$1.7042}
& 0.16
& 0.71 
& 0.136 
& 41.3 \\

\midrule

\multirow{4}{*}{\textbf{IV}}
& Mamba
& 95.75\textsuperscript{\scriptsize $\pm$0.24}
& 0.5148\textsuperscript{\scriptsize $\pm$0.0512}
& 0.7100\textsuperscript{\scriptsize $\pm$0.2913}
& 2.1123\textsuperscript{\scriptsize $\pm$0.6406}
& 0.7051\textsuperscript{\scriptsize $\pm$0.0196}
& 1.42
& 44.67 
& 0.072 
& 32.8 \\

& S7
& 95.28\textsuperscript{\scriptsize $\pm$0.22}
& 0.4475\textsuperscript{\scriptsize $\pm$0.0926}
& 0.5106\textsuperscript{\scriptsize $\pm$0.2924}
& 2.4014\textsuperscript{\scriptsize $\pm$0.8339}
& 0.7255\textsuperscript{\scriptsize $\pm$0.0081}
& 1.38
& 258.58 
& 0.529 
& 116.2 \\

& Jamba
& 95.85\textsuperscript{\scriptsize $\pm$0.23}
& 0.6090\textsuperscript{\scriptsize $\pm$0.0193}
& 0.1792\textsuperscript{\scriptsize $\pm$0.0294}
& 2.9087\textsuperscript{\scriptsize $\pm$0.3797}
& 1.1935\textsuperscript{\scriptsize $\pm$0.0145}
& 0.84
& 6.00 
& 0.069 
& 33.2 \\

& RetNet
& 95.54\textsuperscript{\scriptsize $\pm$0.14}
& 0.6560\textsuperscript{\scriptsize $\pm$0.0272}
& 0.2707\textsuperscript{\scriptsize $\pm$0.0301}
& 3.2325\textsuperscript{\scriptsize $\pm$0.1545}
& 1.1306\textsuperscript{\scriptsize $\pm$0.0940}
& 0.88
& 44.99 
& 0.079 
& 32.8 \\

\midrule

\multirow{5}{*}{\textbf{V}}
& mTAN
& 98.59\textsuperscript{\scriptsize $\pm$0.99}
& 0.5021\textsuperscript{\scriptsize $\pm$0.0112}
& 1.3633\textsuperscript{\scriptsize $\pm$0.1474}
& 3.1877\textsuperscript{\scriptsize $\pm$0.1663}
& 0.0145\textsuperscript{\scriptsize $\pm$0.0044}
& 69.04
& 838.25 
& 0.613 
& 20.5 \\

& ODEFormer
& 98.67\textsuperscript{\scriptsize $\pm$0.73}
& 0.6758\textsuperscript{\scriptsize $\pm$0.0298}
& 0.1359\textsuperscript{\scriptsize $\pm$0.0232}
& 2.7286\textsuperscript{\scriptsize $\pm$0.3995}
& 0.0216\textsuperscript{\scriptsize $\pm$0.0013}
& 46.29
& 259.60 
& 1.011 
& 25.2 \\

& ContiFormer
& 95.79\textsuperscript{\scriptsize $\pm$0.12}
& 0.6096\textsuperscript{\scriptsize $\pm$0.0103}
& 0.1517\textsuperscript{\scriptsize $\pm$0.0188}
& 2.9584\textsuperscript{\scriptsize $\pm$0.2210}
& 0.0334\textsuperscript{\scriptsize $\pm$0.0007}
& 29.90
& 259.72 
& 1.358 
& 29.6 \\

& PDE-Attention
& 95.48\textsuperscript{\scriptsize $\pm$0.17}
& 0.5414\textsuperscript{\scriptsize $\pm$0.0184}
& 0.1388\textsuperscript{\scriptsize $\pm$0.0247}
& 2.5980\textsuperscript{\scriptsize $\pm$0.6489}
& 0.0447\textsuperscript{\scriptsize $\pm$0.0009}
& 22.37
& 258.63 
& 1.692 
& 29.4 \\

& OT-Transformer
& 98.44\textsuperscript{\scriptsize $\pm$0.95}
& 0.6927\textsuperscript{\scriptsize $\pm$0.0133}
& 0.1453\textsuperscript{\scriptsize $\pm$0.0108}
& 2.7302\textsuperscript{\scriptsize $\pm$0.4397}
& 0.0229\textsuperscript{\scriptsize $\pm$0.0074}
& 43.76
& 657.88 
& 7.583 
& 1245.2 \\

\bottomrule
\end{tabular}}
\begin{minipage}{\textwidth}
\footnotesize
\vspace{1mm}
\textbf{Note:}
($\uparrow$) higher is better; ($\downarrow$) lower is better. $^{\dagger}$ Results are reported using a look-back window of 48 and a forecasting horizon of 24. $^{\ddagger}$ Results are reported using a sequence length of 1024 and a batch size of 1.
\vspace{-2mm}
\end{minipage}
\end{table}

\begin{wrapfigure}{r}{0.6\textwidth}
\vspace{-2mm}
\captionsetup{type=table}
\centering
\caption{Software Ecosystem for Continuous-time Machine Learning}
\label{tab:ctml_libraries}
\resizebox{0.58\textwidth}{!}{%
\begin{tabular}{l c c c l}
\toprule
\textbf{Source} & \textbf{PyTorch} & \textbf{JAX} & \textbf{Keras/TF} & \textbf{GitHub Link} \\
\midrule

torchdiffeq & \checkmark & $\times$ & $\times$ & \href{https://github.com/rtqichen/torchdiffeq}{rtqichen/torchdiffeq} \\
        
torchcde & \checkmark & $\times$ & $\times$ & \href{https://github.com/patrick-kidger/torchcde}{patrick-kidger/torchcde} \\
        
torchsde & \checkmark & $\times$ & $\times$ & \href{https://github.com/google-research/torchsde}{google-research/torchsde} \\
        
TorchDyn & \checkmark & $\times$ & $\times$ & \href{https://github.com/DiffEqML/torchdyn}{DiffEqML/torchdyn} \\
        
Diffrax & $\times$ & \checkmark & $\times$ & \href{https://github.com/patrick-kidger/diffrax}{patrick-kidger/diffrax} \\
        
diffeqpy & \checkmark & \checkmark & $\times$ & \href{https://github.com/SciML/diffeqpy}{SciML/diffeqpy} \\
        
Mamba & \checkmark & $\times$ & $\times$ & \href{https://github.com/state-spaces/mamba}{state-spaces/mamba} \\
        
mamba-py & \checkmark & $\times$ & $\times$ & \href{https://github.com/alxndrTL/mamba.py}{alxndrTL/mamba.py} \\
        
mamba-jax & $\times$ & \checkmark & $\times$ & \href{https://github.com/walrus-ai/mamba-jax}{walrus-ai/mamba-jax} \\
        
S4 & \checkmark & $\times$ & $\times$ & \href{https://github.com/state-spaces/s4}{state-spaces/s4} \\
        
S5 & $\times$ & \checkmark & $\times$ & \href{https://github.com/lindermanlab/S5}{lindermanlab/S5} \\
        
LRU & \checkmark & \checkmark & $\times$ & \href{https://github.com/NicolasZucchet/minimal-LRU}{NicolasZucchet/minimal-LRU} \\
        
ncps (LTC / CfC) & \checkmark & $\times$ & \checkmark & \href{https://github.com/mlech26l/ncps}{mlech2615/ncps} \\
        
LAN/NAC/NSAC & \checkmark & $\times$ & \checkmark & \href{https://github.com/itxwaleedrazzaq/keras-nac}{itxwaleedrazzaq/keras-nac}/ \href{https://github.com/itxwaleedrazzaq/torch-nac}{torch-nac} \\
        
ContiFormer & \checkmark & $\times$ & $\times$ & \href{https://github.com/microsoft/SeqML/tree/main/ContiFormer}{microsoft/ContiFormer} \\

\bottomrule
\end{tabular}}
\vspace{-2mm}
\end{wrapfigure}

\textbf{\textit{Interpretation:}} Table~\ref{tab:all_results} suggests that no single family dominates across all tasks, broadly consistent with the design trade-offs implied by the canonical formulation. On irregular time series, \emph{Family II} appears the strongest: PhasedLSTM attains the best E-MNIST accuracy (98.72\%), CT-GRU the best PhysioNet AUC (0.837), with PhasedLSTM and FLUID also above 0.79. The input-dependent time constant mechanism allows these models to weight observations by arrival time.  \emph{Family V} is close behind on E-MNIST (ODEFormer 98.67\%, mTAN 98.59\%) but weaker on PhysioNet (0.50-0.69), which indicates that time-aware attention is insufficient when query construction is not itself learned. On long-range forecasting, \emph{Family I} leads ETTm1 (HiPPO 0.054, S4 0.066), while Jena-Climate is led by CT-RNN (0.9270), followed by DeepState (1.189); \emph{Family V} trails there (2.60-3.19), and \emph{Family IV} is the weakest. This pattern is consistent with the hypothesis that a fixed structured basis favors long-horizon compression, whereas input-dependent parametrization favors content selectivity over compressed history. Runtime exhibits a formulation-driven split: parallel architectures (S4 0.022, mTAN 0.015, ODEFormer 0.022, OT-Transformer 0.023) run two to three orders of magnitude faster than the solver-based architectures of \emph{Family II \& III} (NSDE 13.8s, LTC 12.7s, CT-RNN 12.0s), with \emph{Family IV} intermediate (0.7-1.2). Memory follows the inverse pattern of runtime, peaking at 259–838 MB for attention-based models (mTAN, OT-Transformer) against under 3~MB for recurrent formulations. Overall, improved accuracy of expressive architectures comes at substantial computational cost, while structured linear and selective-scan formulations favor efficient deployment with weaker accuracy.
\subsection{Ecosystems}
The software ecosystem is fragmented across multiple frameworks, predominantly PyTorch, but also including JAX and TensorFlow/Keras. For \emph{Family III}, comprehensive ODE solver wrappers with adjoint sensitivity support are available in PyTorch (via \texttt{torchdiffeq}, \texttt{torchdyn}, and \texttt{torchsde}) and JAX (via \texttt{diffrax}). Implementations for \emph{Family I \& II} are similarly distributed: the original S4 and Mamba models use PyTorch with custom CUDA kernels to achieve near-optimal memory throughput (with \texttt{mamba-py} providing a Python interface), whereas the S5 model uses JAX with parallel scans. Within \emph{Family II}, the NCPs library implements CfC and LTC networks in PyTorch and TensorFlow, while \texttt{keras-nac} and \texttt{torch-nac} offer TensorFlow and PyTorch implementations of the LAN, NAC, and NSAC models, respectively. Finally, \emph{Family V} lacks a unified library, instead relying on individual GitHub repositories. We summarize the available implementations and their corresponding frameworks in Table~\ref{tab:ctml_libraries}.

\subsection*{Key Takeaways}
The benchmark analysis suggests that no single family dominates across all four tasks: \emph{Family II} attains the best irregular-time results (PhasedLSTM and CT-GRU lead E-MNIST and PhysioNet); \emph{Family I} leads ETTm1 forecasting (HiPPO, S4); and \emph{Family II} leads Jena-Climate (CT-RNN); \emph{Family IV} is the weakest family for forecasting on ETTm1 but is mid-pack on E-MNIST; \emph{Family V} is competitive on E-MNIST but trails on both forecasting columns. We hypothesize these patterns track each family’s core mechanism: adaptive time constants and attention over continuous trajectories favor sparse, event-like data; structured linear memory suits smooth, stationary sequences, and input-selective discretization offers little benefit at small scale on short, low-dimensional tasks. The runtime results reinforce the expressiveness-efficiency trade-off identified throughout this survey: solver-based \emph{Family II \& III} are the slowest, \emph{Family V} is the most memory hungry, and \emph{Family I \& IV} are the most efficient to run. These results come from a single untuned configuration and are best treated as illustrative rather than definitive. The software ecosystem for CT families remains fragmented across frameworks, with PyTorch being the dominant platform, and a unified library remains an open challenge.

\section{Open Problems}\label{section:open_problem}
Despite the progress surveyed in the preceding sections, fundamental challenges remain across every aspect of CT machine learning. Below, we highlight the most significant of these open problems, grouped by their nature and tractability.

\subsection{Theoretical Challenges}

\textbf{\textit{Approximation Theory:}} Universal approximation guarantees exist only partially for \emph{Family II \& III}~\cite{Funahashi1993, hasani2021liquid, razzaq2025neuronal, kidger2020neural}, and only for smooth flows over compact time intervals. No general theory of approximation rates exists; therefore, it remains unclear how model capacity should scale to approximate a target dynamical system to a given accuracy, or how model width, depth, temporal horizon, and approximation error relate to one another. The spectral bias of CT models and their tendency to learn low-frequency components before high-frequency ones are also less understood than the neural-tangent-kernel (NTK) theory for feedforward networks~\cite{jacot2018neural}.

\textbf{\textit{Stability and Robustness:}} The adjoint sensitivity method used to train \emph{Family III} introduces approximation errors, and how these errors affect gradient accuracy has not been formally analyzed. The interaction between ODE stiffness, step-size selection, and gradient quality is understood mainly through empirical evidence rather than theory. In \emph{Family I \& IV}, HiPPO initialization gives stable eigenvalues at the start of training, but optimization can push these eigenvalues outside the stability region, and we do not yet know under what conditions this happens. Stability analysis of \emph{Family V} is underdeveloped with no established theory of how dynamics attention-based behave under perturbations.

\subsection{Algorithmic and Computational Challenges}

\textit{\textbf{Expressivity vs. Efficiency:}} Throughout this survey, we have framed the central challenge as balancing expressivity against computational efficiency. Rigorous, empirically validated characterizations of where different model families sit along this spectrum are still missing. CT-RNNs with $\tanh$ activation and Mamba’s selective scan share the same state-space formulation but differ by orders of magnitude in practical efficiency. NDEs with adaptive solvers can require thousands of function evaluations for stiff systems, whereas fixed-step solvers trade accuracy for speed. Existing literature reports these trade-offs with inconsistent evaluation protocols and metrics; a systematic comparison across families remains difficult.

\textbf{\textit{Training instability in Family III:}} The adjoint method reduces memory usage to $\mathcal{O}(1)$ with respect to $S$, but this advantage often comes at the cost of numerical instability, especially for stiff or chaotic systems. Backpropagating directly through the solver avoids this issue but requires storing every intermediate state, increasing memory complexity to $\mathcal{O}(S)$, which becomes impractical for long sequences. While Gholaminejad et al.~\cite{ijcai2019p103} and Rackauckas et al.~\cite{rackauckas2020universal} compared these approaches empirically, no broadly accepted solution has emerged.  Combining the memory efficiency of the adjoint method with the accuracy of BPTT, whether through checkpointing, reversible solver formulations, or some other approach, remains an important open problem.

\textbf{\textit{Selective Scan Generality:}} Mamba’s parallel associative scan is highly efficient for scalar or diagonal state transitions but does not naturally extend to full matrix-valued transitions. The SSD framework introduced in Mamba-2 further restricts $A$ to scalar values to maximize hardware efficiency. Whether selective scan algorithms can support richer state transitions while keeping linear computational complexity will shape how SSMs develop from here.

\textbf{\textit{Unified Benchmark Suite:}} 
% As discussed in Section~\ref{section:training}, 
The absence of a unified benchmark suite with standardized compute budgets makes even basic empirical comparisons across model families difficult. Of all the open problems, this is among the most tractable: a coordinated benchmarking effort would give the field a much clearer basis for measuring real progress.

\subsection{Scientific and Methodological Challenges}

% \textbf{\textit{Revisiting Irregular Sampling:}}  A recurring claim in the CT literature is that these models outperform discrete approaches on irregularly sampled time series. The evidence supports this for clinical benchmarks, where CT methods consistently achieve AUC improvements of roughly 4-10\% over discrete baselines. But the claim is often generalized to regularly sampled data as well, where the CT formulation may offer little intrinsic benefit. Much of the SSM forecasting literature, which dominates current time-series benchmarks, focuses almost exclusively on regularly sampled datasets, so the distinction between a genuine advantage for irregular sampling and merely efficient modeling of regular sequences is easy to miss.

\textbf{\textit{Discretization Analysis:}} Discretization provides the link between CT theory and discrete computation. Across all families, the choice of discretization scheme influences both numerical stability and approximation accuracy. For example, in \textit{HiPPO-LegS} bilinear discretization preserves the normal matrix property, whereas Euler discretization does not; in Mamba, the discretization step is itself an input-dependent gate that sets the effective temporal resolution. Despite its central role, the impact of discretization on model behavior has received little systematic study. A general theoretical framework explaining how discretization interacts with system dynamics, sampling pattern, and training objectives remains an open problem.

\textbf{\textit{Foundation Model Challenges:}} Pretraining strategies for CT models are in their early stages. The success of self-supervised pretraining in language models and vision transformers has motivated similar approaches for time-series foundation models, including Mamba4Cast~\cite{bhethanabhotla2024mamba4cast}, TimesFM~\cite{das2024decoderonly}, and Lag-Llama~\cite{rasul2024lagllama}, all pretrained on large collections of heterogeneous time series and then adapted to downstream tasks. However, the pretraining design space for time series is substantially more complex than for language or vision. Should pretraining rely on irregularly or regularly sampled data? Which objectives work best, and how should variation in observed variables and sampling patterns be handled? No consensus has emerged on any of these questions.

\textit{\textbf{Limited Scaling:}} The relationship between model scale, dataset size, and performance has been extensively studied for language models, but comparable analyses are largely absent. It remains unclear whether CT models follow scaling trends similar to the DT models or whether their formulation introduces different scaling behavior. Understanding this relationship is important for guiding resource allocation and model development strategies for future foundation models.

\subsection{Interdisciplinary Challenges}

\textit{\textbf{Interpretability:}} Gradient-based attribution techniques developed for DT-transformers do not directly extend to CT architectures. CT models encode information through hidden state trajectories rather than discrete representations, which raises new challenges for visualization and attribution. Parameters such as learned time constant and discretization steps strongly influence model dynamics but have no direct counterparts in standard discrete architectures, and how they contribute to prediction behavior remains poorly understood.

\textbf{\textit{Beyond Data-Driven Learning:}} CT models are particularly well-suited to scientific applications where prior knowledge of system dynamics is available in the form of differential equations. Physics-informed neural networks (PINNs)~\cite{raissi2019physics} and Neural Operators~\cite{li2020fourier} already incorporate such constraints, but they have largely developed as a separate line of work from the area covered in this survey. Combining symbolic domain knowledge with learned CT dynamics remains a largely unexplored research direction.

\section{Research Agenda}\label{section:agenda}

The open problems discussed in Section~\ref{section:open_problem} motivate the research agenda below, which we organize by expected time horizon and potential impact. 

\subsection{Near-Term Agenda (2--3 years)}
\textit{\textbf{Developing a Unified Benchmark Suite:}} An important near-term goal is to develop a benchmark suite that evaluates all families on a common set of datasets under controlled computational budgets. This suite should include at least five datasets from distinct domains such as clinical monitoring, environmental sensing, financial markets, industrial process control, and physical system modeling. It should use consistent preprocessing and data splits, control for model size and training budgets, and report prediction accuracy, calibration, computational cost, data efficiency, and inference time. Consistent reporting of variance and sufficient information for reproducibility should also be part of the benchmark protocol.

\textit{\textbf{Analysis of Discretizations:}} The choice of discretization operator influences numerical stability, approximation accuracy, and long-term dynamics. However, the effects of discretization choices across families have only been characterized through empirical evaluation rather than theoretical analysis. Controlled theoretical studies of the interaction between discretization strategy, system dynamics, sampling schedule, and training objective could provide useful practical guidance and potentially reveal connections between seemingly different CT models.

\subsection{Mid-Term Agenda (4--5 years)}

\textit{\textbf{Optimization for stiffness:}} Training \emph{Family III} architectures on stiff or chaotic systems remains challenging. Two directions appear particularly important: (a) numerical methods that retain the memory advantages of the adjoint method while producing more accurate gradients, and (b) ODE solvers that support efficient checkpointing, combined with regularization or parameterization strategies that improve training stability in stiff and chaotic regimes.

\textit{\textbf{Scalability of State Transitions:}} Existing parallel associative-scan algorithms are mostly limited to scalar or diagonal state transitions. Extending them to structured matrices, including low-rank, banded, and sparse form, without losing near-linear computational complexity could improve the expressiveness of SSMs while preserving their efficiency. Advances in structured matrix decomposition and hardware kernel fusion are promising, but existing approaches have yet to match the scalability and selectivity achieved with scalar or diagonal state transitions.

\subsection{Long-Term Agenda (> 5 years)}

\textit{\textbf{Pre-trained Foundation Models:}} Time-series foundation models such as Mamba4Cast and TimesFM raise a broader question: can large-scale pretraining on diverse time-series data provide benefits similar to those seen in language and vision? Three directions appear particularly important: pretraining objectives that capture temporal structure without task-specific supervision, architectures that can accommodate heterogeneous variables and sampling , and scaling laws that connect model size and data diversity with downstream performance.

\textit{\textbf{Hardware-Aware Development:}} Hardware fragmentation remains a practical challenge. Specialized platforms that efficiently support multiple CT computational frameworks, including ODE solvers, linear recurrence, and selective-scan algorithms, could enable more standard evaluation and facilitate the transition from research prototypes to practical systems. Analog and mixed-signal processors, which can perform CT computations natively, are promising direction, although their maturity and programmability still need substantial improvement.

\section{Conclusion}\label{section:conclusion}
In this survey, we have argued that continuous-time (CT) machine learning is best understood not as competing paradigms, but as a unified design space. We introduced a concept-driven taxonomy that organizes all major branches according to their underlying base functional form into five families: (i) \textit{Linear Dynamical Systems}; (ii) \textit{Linearly-Coupled Dynamics with State-Dependent Decay}; (iii) \textit{Freely Parameterized Vector Fields}; (iv) \textit{Selective Scan State-Space Models}; and (v) \textit{Continuous-Time Transformers}. We described that these families can be interpreted as instantiations of a canonical dynamical system with different architectural choices imposed on the governing vector field, the treatment of stochasticity, the memory mechanism, and discretization.\\
For practitioners, this taxonomy frames model selection as choosing the level of inductive structure appropriate for the problem: \emph{Family I \& IV} favor efficient parallel training and constant-memory inference at some cost in expressiveness; \emph{Family II} trades some parallel efficiency for adaptable, interpretable dynamics; \emph{Family III} maximize expressiveness at the cost of solver-dependent compute; and \emph{Family V} pays quadratic attention cost for flexibility over continuous trajectories. The theoretical complexity and empirical analyses quantify these trade-offs in illustrative architecture-controlled settings. Furthermore, robustness to irregular sampling and computational efficiency should be viewed as consequences of specific design choices rather than inherent properties of any family.\\
Several challenges remain. Approximation theory covers only a subset of CT models, while the interplay among discretization, sampling patterns, and optimization is insufficiently understood. Fragmented benchmarks, inconsistent evaluation protocols, and the lack of scaling laws continue to hinder rigorous comparison across families.\\
Future progress will require standardized benchmarks, a stronger theoretical account of discretization, and scalable training methods for increasingly expressive CT architectures. Extending these models toward foundation models, scientific machine learning, and hardware-aware implementation offers a promising direction for establishing CT learning as a unified framework for temporal representation learning.

\section*{Data and Code Availability} 
The code is publicly available at \url{https://github.com/itxwaleedrazzaq/ctml-review}.

% \section*{Acknowledgments}
% This research was supported by the CAS-ANSO Scholarship. We acknowledge the intellectual and material contributions of the University of Science and Technology of China (USTC) and the Alliance of International Science Organizations (ANSO). AI/LLM tools were used to polish the writing of the manuscript under strict human supervision.

% \section*{Authors Contribution}
% \textbf{Waleed Razzaq:} Conceptualization, Methodology, Data Curation, Writing- Original draft preparation. \textbf{Yun-Sheng Zhao}: Writing- Reviewing.  \textbf{Yun-Bo Zhao}: Supervision, Writing- Reviewing. 

% The authors declare that they have no known competing financial interests or personal relationships that could have appeared to influence the work reported in this paper.

% \section*{Ethics Approval}
% This study was conducted in accordance with ethical standards. 

\bibliographystyle{unsrt}
\bibliography{references}

@article{chen2018neural,
  title={Neural ordinary differential equations},
  author={Chen, Ricky TQ and Rubanova, Yulia and Bettencourt, Jesse and Duvenaud, David K},
  journal={Advances in neural information processing systems},
  volume={31},
  year={2018}
}

@inproceedings{ansari2023neural,
  title={Neural continuous-discrete state space models for irregularly-sampled time series},
  author={Ansari, Abdul Fatir and Heng, Alvin and Lim, Andre and Soh, Harold},
  booktitle={International Conference on Machine Learning},
  pages={926--951},
  year={2023},
  organization={PMLR}
}

@inproceedings{hasani2021liquid,
  title={Liquid time-constant networks},
  author={Hasani, Ramin and Lechner, Mathias and Amini, Alexander and Rus, Daniela and Grosu, Radu},
  booktitle={Proceedings of the AAAI conference on artificial intelligence},
  volume={35},
  number={9},
  pages={7657--7666},
  year={2021}
}

@article{hasani2018liquid,
  title={Liquid time-constant recurrent neural networks as universal approximators},
  author={Hasani, Ramin M and Lechner, Mathias and Amini, Alexander and Rus, Daniela and Grosu, Radu},
  journal={arXiv preprint arXiv:1811.00321},
  year={2018}
}

@article{hasani2022closed,
  title={Closed-form continuous-time neural networks},
  author={Hasani, Ramin and Lechner, Mathias and Amini, Alexander and Liebenwein, Lucas and Ray, Aaron and Tschaikowski, Max and Teschl, Gerald and Rus, Daniela},
  journal={Nature Machine Intelligence},
  volume={4},
  number={11},
  pages={992--1003},
  year={2022},
  publisher={Nature Publishing Group UK London}
}

@article{liang2026rederived,
  title={Rederived Closed-Form Continuous-Time Neural Networks},
  author={Liang, Xingyu and Zhou, Siyi and Chen, Li'Ao and Weng, Liguo and Lin, Haifeng and Huang, Yuxin and Xia, Min},
  journal={IEEE Transactions on Neural Networks and Learning Systems},
  year={2026},
  publisher={IEEE}
}

@article{lechner2018neuronal,
  title={Neuronal circuit policies},
  author={Lechner, Mathias and Hasani, Ramin M and Grosu, Radu},
  journal={arXiv preprint arXiv:1803.08554},
  year={2018}
}

@inproceedings{chien2021continuous,
  title={Continuous-time attention for sequential learning},
  author={Chien, Jen-Tzung and Chen, Yi-Hsiang},
  booktitle={Proceedings of the AAAI conference on artificial intelligence},
  volume={35},
  number={8},
  pages={7116--7124},
  year={2021}
}

@article{chen2023contiformer,
  title={Contiformer: Continuous-time transformer for irregular time series modeling},
  author={Chen, Yuqi and Ren, Kan and Wang, Yansen and Fang, Yuchen and Sun, Weiwei and Li, Dongsheng},
  journal={Advances in Neural Information Processing Systems},
  volume={36},
  pages={47143--47175},
  year={2023}
}

@article{d2023odeformer,
  title={Odeformer: Symbolic regression of dynamical systems with transformers},
  author={d'Ascoli, St{\'e}phane and Becker, S{\"o}ren and Mathis, Alexander and Schwaller, Philippe and Kilbertus, Niki},
  journal={arXiv preprint arXiv:2310.05573},
  year={2023}
}

@article{kan2025ot,
  title={OT-Transformer: a continuous-time transformer architecture with optimal transport regularization},
  author={Kan, Kelvin and Li, Xingjian and Osher, Stanley},
  journal={arXiv preprint arXiv:2501.18793},
  year={2025}
}

@article{razzaq2025neuronal,
title = {Neuronal Attention Circuit (NAC) for Representation Learning},
journal = {Neural Networks},
pages = {109495},
year = {2026},
issn = {0893-6080},
doi = {https://doi.org/10.1016/j.neunet.2026.109495},
author = {Waleed Razzaq and Yun-Bo Zhao},
}

@article{razzaq2026fluid,
  title={FLUID: Continuous-Time Hyperconnected Sparse Transformer for Sink-Free Learning},
  author={Razzaq, Waleed and Zhao, Yun-Bo},
  journal={arXiv preprint arXiv:2605.04421},
  year={2026}
}

@article{razzaq2026neuronal,
  title={Neuronal Stochastic Attention Circuit (NSAC) for Probabilistic Representation Learning},
  author={Razzaq, Waleed and Zhao, Yun-Bo},
  journal={arXiv preprint arXiv:2605.26061},
  year={2026}
}

@article{kidger2022neural,
  title={On neural differential equations},
  author={Kidger, Patrick},
  journal={arXiv preprint arXiv:2202.02435},
  year={2022}
}

@article{oh2025comprehensive,
  title={Comprehensive review of neural differential equations for time series analysis},
  author={Oh, YongKyung and Kam, Seungsu and Lee, Jonghun and Lim, Dong-Young and Kim, Sungil and Bui, Alex},
  journal={arXiv preprint arXiv:2502.09885},
  year={2025}
}

@article{somvanshi2025s4,
  title={From s4 to mamba: A comprehensive survey on structured state space models},
  author={Somvanshi, Shriyank and Islam, Md Monzurul and Mimi, Mahmuda Sultana and Polock, Sazzad Bin Bashar and Chhetri, Gaurab and Das, Subasish},
  journal={arXiv preprint arXiv:2503.18970},
  pages={1--30},
  year={2025}
}

@article{tiezzi2025state,
  title={State-space modeling in long sequence processing: A survey on recurrence in the transformer era},
  author={Tiezzi, Matteo and Casoni, Michele and Betti, Alessandro and Gori, Marco and Melacci, Stefano},
  journal={Neural Networks},
  pages={108039},
  year={2025},
  publisher={Elsevier}
}

@article{wen2022transformers,
  title={Transformers in time series: A survey},
  author={Wen, Qingsong and Zhou, Tian and Zhang, Chaoli and Chen, Weiqi and Ma, Ziqing and Yan, Junchi and Sun, Liang},
  journal={arXiv preprint arXiv:2202.07125},
  year={2022}
}

@inproceedings{sommers2024survey,
  title={A survey of transformer enabled time series synthesis},
  author={Sommers, Alexander and Cummins, Logan and Mittal, Sudip and Rahimi, Shahram and Seale, Maria and Jaboure, Joseph and Arnold, Thomas},
  booktitle={2024 IEEE 10th International Conference on Collaboration and Internet Computing (CIC)},
  pages={60--69},
  year={2024},
  organization={IEEE}
}

@inproceedings{katharopoulos2020transformers,
  title={Transformers are rnns: Fast autoregressive transformers with linear attention},
  author={Katharopoulos, Angelos and Vyas, Apoorv and Pappas, Nikolaos and Fleuret, Fran{\c{c}}ois},
  booktitle={International conference on machine learning},
  pages={5156--5165},
  year={2020},
  organization={PMLR}
}

@article{dao2024transformers,
  title={Transformers are ssms: Generalized models and efficient algorithms through structured state space duality},
  author={Dao, Tri and Gu, Albert},
  journal={arXiv preprint arXiv:2405.21060},
  year={2024}
}

@article{vaswani2017attention,
  title={Attention is all you need},
  author={Vaswani, Ashish and Shazeer, Noam and Parmar, Niki and Uszkoreit, Jakob and Jones, Llion and Gomez, Aidan N and Kaiser, {\L}ukasz and Polosukhin, Illia},
  journal={Advances in neural information processing systems},
  volume={30},
  year={2017}
}

@article{rangapuram2018deep,
  title={Deep state space models for time series forecasting},
  author={Rangapuram, Syama Sundar and Seeger, Matthias W and Gasthaus, Jan and Stella, Lorenzo and Wang, Yuyang and Januschowski, Tim},
  journal={Advances in neural information processing systems},
  volume={31},
  year={2018}
}

@article{hochreiter1997long,
  title={Long short-term memory},
  author={Hochreiter, Sepp and Schmidhuber, J{\"u}rgen},
  journal={Neural computation},
  volume={9},
  number={8},
  pages={1735--1780},
  year={1997},
  publisher={MIT press}
}

@article{cho2014learning,
  title={Learning phrase representations using RNN encoder-decoder for statistical machine translation},
  author={Cho, Kyunghyun and Van Merri{\"e}nboer, Bart and Gulcehre, Caglar and Bahdanau, Dzmitry and Bougares, Fethi and Schwenk, Holger and Bengio, Yoshua},
  journal={arXiv preprint arXiv:1406.1078},
  year={2014}
}

@article{mcculloch1943logical,
  title={A logical calculus of the ideas immanent in nervous activity},
  author={McCulloch, Warren S and Pitts, Walter},
  journal={The bulletin of mathematical biophysics},
  volume={5},
  number={4},
  pages={115--133},
  year={1943},
  publisher={Springer}
}

@inproceedings{finlay2020train,
  title={How to train your neural {ODE}: The world of {J}acobian and kinetic regularization},
  author={Finlay, Chris and Jacobsen, Joern-Henrik and Nurbekyan, Levon and Oberman, Adam},
  booktitle={Proceedings of the 37th International Conference on Machine Learning},
  pages={3154--3164},
  year={2020},
  volume={119},
  series={Proceedings of Machine Learning Research},
  publisher={PMLR}
}

@inproceedings{kelly2020learning,
  title={Learning Differential Equations that are Easy to Solve},
  author={Kelly, Jacob and Bettencourt, Jesse and Johnson, Matthew James and Duvenaud, David},
  booktitle={Advances in Neural Information Processing Systems},
  volume={33},
  pages={5739--5750},
  year={2020}
}

@article{hodgkin1952quantitative,
  title={A quantitative description of membrane current and its application to conduction and excitation in nerve},
  author={Hodgkin, Alan L and Huxley, Andrew F},
  journal={The Journal of physiology},
  volume={117},
  number={4},
  pages={500},
  year={1952}
}

@article{Hopfield1984,
  author    = {Hopfield, J. J.},
  title     = {Neurons with graded response have collective computational properties like those of two-state neurons},
  journal   = {Proceedings of the National Academy of Sciences},
  volume    = {81},
  number    = {10},
  pages     = {3088--3092},
  year      = {1984},
  publisher = {National Academy of Sciences},
  doi       = {10.1073/pnas.81.10.3088},
  url       = {https://www.pnas.org/doi/10.1073/pnas.81.10.3088}
}

@article{hopfield1985neural,
  title={“Neural” computation of decisions in optimization problems},
  author={Hopfield, John J and Tank, David W},
  journal={Biological cybernetics},
  volume={52},
  number={3},
  pages={141--152},
  year={1985},
  publisher={Springer}
}

@article{funahashi1989approximate,
  title={On the approximate realization of continuous mappings by neural networks},
  author={Funahashi, Ken-Ichi},
  journal={Neural networks},
  volume={2},
  number={3},
  pages={183--192},
  year={1989},
  publisher={Elsevier}
}

@article{Funahashi1993,
  title     = {Approximation of dynamical systems by continuous time recurrent neural networks},
  author    = {Ken-ichi Funahashi and Yuichi Nakamura},
  journal   = {Neural Networks},
  volume    = {6},
  number    = {6},
  pages     = {801--806},
  year      = {1993},
  publisher = {Elsevier},
  doi       = {10.1016/S0893-6080(05)80125-X}
}

@article{cybenko1989approximation,
  title={Approximation by superpositions of a sigmoidal function},
  author={Cybenko, George},
  journal={Mathematics of control, signals and systems},
  volume={2},
  number={4},
  pages={303--314},
  year={1989},
  publisher={Springer}
}

@article{lecun1998gradient,
  title={Gradient-based learning applied to document recognition},
  author={LeCun, Yann and Bottou, L{\'e}on and Bengio, Yoshua and Haffner, Patrick},
  journal={Proceedings of the IEEE},
  volume={86},
  number={11},
  pages={2278--2324},
  year={1998},
  publisher={Ieee}
}

@article{beer1995dynamics,
  title={On the dynamics of small continuous-time recurrent neural networks},
  author={Beer, Randall D},
  journal={Adaptive Behavior},
  volume={3},
  number={4},
  pages={469--509},
  year={1995},
  publisher={Sage Publications Sage CA: Thousand Oaks, CA}
}

@article{elman1990finding,
  title={Finding structure in time},
  author={Elman, Jeffrey L},
  journal={Cognitive science},
  volume={14},
  number={2},
  pages={179--211},
  year={1990},
  publisher={Wiley Online Library}
}

@article{rumelhart1986learning,
  title={Learning representations by back-propagating errors},
  author={Rumelhart, David E and Hinton, Geoffrey E and Williams, Ronald J},
  journal={nature},
  volume={323},
  number={6088},
  pages={533--536},
  year={1986},
  publisher={Nature Publishing Group UK London}
}

@article{rubanova2019latent,
  title={Latent ordinary differential equations for irregularly-sampled time series},
  author={Rubanova, Yulia and Chen, Ricky TQ and Duvenaud, David K},
  journal={Advances in neural information processing systems},
  volume={32},
  year={2019}
}

@article{lechner2022mixed,
  title={Mixed-memory rnns for learning long-term dependencies in irregularly sampled time series},
  author={Lechner, Mathias and Hasani, Ramin},
  year={2022}
}

@article{kidger2020neural,
  title={Neural controlled differential equations for irregular time series},
  author={Kidger, Patrick and Morrill, James and Foster, James and Lyons, Terry},
  journal={Advances in neural information processing systems},
  volume={33},
  pages={6696--6707},
  year={2020}
}

@inproceedings{li2020scalable,
  title={Scalable gradients for stochastic differential equations},
  author={Li, Xuechen and Wong, Ting-Kam Leonard and Chen, Ricky TQ and Duvenaud, David},
  booktitle={International conference on artificial intelligence and statistics},
  pages={3870--3882},
  year={2020},
  organization={PMLR}
}

@article{kong2020sde,
  title={Sde-net: Equipping deep neural networks with uncertainty estimates},
  author={Kong, Lingkai and Sun, Jimeng and Zhang, Chao},
  journal={arXiv preprint arXiv:2008.10546},
  year={2020}
}

@article{voelker2019legendre,
  title={Legendre memory units: Continuous-time representation in recurrent neural networks},
  author={Voelker, Aaron and Kaji{\'c}, Ivana and Eliasmith, Chris},
  journal={Advances in neural information processing systems},
  volume={32},
  year={2019}
}

@article{gu2021efficiently,
  title={Efficiently modeling long sequences with structured state spaces},
  author={Gu, Albert and Goel, Karan and R{\'e}, Christopher},
  journal={arXiv preprint arXiv:2111.00396},
  year={2021}
}

@article{amini2025lfm2,
  title={Lfm2 technical report},
  author={Amini, Alexander and Banaszak, Anna and Benoit, Harold and B{\"o}{\"o}k, Arthur and Dakhran, Tarek and Duong, Song and Eng, Alfred and Fernandes, Fernando and H{\"a}rk{\"o}nen, Marc and Harrington, Anne and others},
  journal={arXiv preprint arXiv:2511.23404},
  year={2025}
}

@article{jhin2024attentive,
  title={Attentive neural controlled differential equations for time-series classification and forecasting},
  author={Jhin, Sheo Yon and Shin, Heejoo and Kim, Sujie and Hong, Seoyoung and Jo, Minju and Park, Solhee and Park, Noseong and Lee, Seungbeom and Maeng, Hwiyoung and Jeon, Seungmin},
  journal={Knowledge and Information Systems},
  volume={66},
  number={3},
  pages={1885--1915},
  year={2024},
  publisher={Springer}
}

@inproceedings{jhin2021ace,
  title={Ace-node: Attentive co-evolving neural ordinary differential equations},
  author={Jhin, Sheo Yon and Jo, Minju and Kong, Taeyong and Jeon, Jinsung and Park, Noseong},
  booktitle={Proceedings of the 27th ACM SIGKDD Conference on Knowledge Discovery \& Data Mining},
  pages={736--745},
  year={2021}
}

@inproceedings{zhang2025continuous,
  title={Continuous-Time Attention: PDE-Guided Mechanisms for Long-Sequence Transformers},
  author={Zhang, Yukun and Zhou, Xueqing},
  booktitle={Proceedings of the 2025 Conference on Empirical Methods in Natural Language Processing},
  pages={21654--21674},
  year={2025}
}

@article{shukla2021multi,
  title={Multi-time attention networks for irregularly sampled time series},
  author={Shukla, Satya Narayan and Marlin, Benjamin M},
  journal={arXiv preprint arXiv:2101.10318},
  year={2021}
}

@article{hasani2022liquid,
  title={Liquid structural state-space models},
  author={Hasani, Ramin and Lechner, Mathias and Wang, Tsun-Hsuan and Chahine, Makram and Amini, Alexander and Rus, Daniela},
  journal={arXiv preprint arXiv:2209.12951},
  year={2022}
}

@article{cantini2025exact,
  title={Exact implementation of closed-form liquid neural networks with arbitrary precision},
  author={Cantini, Clotilde and Rolland-Piegue, Emilie and Schmitter, Daniel},
  journal={IEEE Signal Processing Letters},
  volume={32},
  pages={921--925},
  year={2025},
  publisher={IEEE}
}

@article{smith2208simplified,
  title={Simplified state space layers for sequence modeling},
  author={Smith, Jimmy TH and Warrington, Andrew and Linderman, Scott W},
  journal={arXiv preprint arXiv:2208.04933},
  year={2022}
}

@book{wanner1996solving,
  title={Solving ordinary differential equations II},
  author={Wanner, Gerhard and Hairer, Ernst},
  volume={375},
  year={1996},
  publisher={Springer Berlin Heidelberg New York}
}

@inproceedings{gu2020hippo,
  author    = {Gu, Albert and Dao, Tri and Ermon, Stefano and Atzmon, Fuad and R\'{e}, Christopher},
  booktitle = {Advances in Neural Information Processing Systems},
  editor    = {H. Larochelle and M. Ranzato and R. Hadsell and M. F. Balcan and H. Lin},
  pages     = {1474--1487},
  publisher = {Curran Associates, Inc.},
  title     = {HiPPO: Recurrent Memory with Optimal Polynomial Projections},
  url       = {https://proceedings.neurips.cc/paper/2020/file/102f0bb6efb3a6128a3c750dd16729be-Paper.pdf},
  volume    = {33},
  year      = {2020}}

@inproceedings{gu2021combining,
  title={Combining Recurrent, Convolutional, and Continuous-time Models via Linear State Space Layer},
  author={Gu, Albert and Johnson, Isys and Goel, Karan and Saab, Khaled and Dao, Tri and Rudra, Atri and R{\'e}, Christopher},
  booktitle={Advances in Neural Information Processing Systems},
  volume={34},
  pages={23901--23912},
  year={2021}
}

@inproceedings{gupta2022diagonal,
  title={Diagonal State Spaces are as Effective as Structured State Spaces},
  author={Gupta, Ankit and Gu, Albert and Berant, Jonathan},
  booktitle={Advances in Neural Information Processing Systems},
  volume={35},
  pages={22982--22994},
  year={2022}
}

@inproceedings{gu2022parameterization,
  title={On the Parameterization and Initialization of Diagonal State Space Models},
  author={Gu, Albert and Gupta, Ankit and Goel, Karan and R{\'e}, Christopher},
  booktitle={Advances in Neural Information Processing Systems},
  volume={35},
  pages={35971--35985},
  year={2022}
}

@inproceedings{orvieto2023resurrecting,
  title={Resurrecting Recurrent Neural Networks for Long Sequences},
  author={Orvieto, Antonio and Smith, Samuel L and Gu, Albert and Anagnostidis, Sotiris and Hofmann, Thomas and De, Soham},
  booktitle={International Conference on Machine Learning (ICML)},
  pages={26670--26692},
  year={2023}
}

@inproceedings{mehta2022long,
  title={Long Range Language Modeling via Gated State Spaces},
  author={Mehta, Harsh and Gupta, Ankit and Cutkosky, Ashok and Neyshabur, Behnam},
  booktitle={International Conference on Learning Representations (ICLR)},
  year={2023}
}

@article{cohen1983absolute,
  title={Absolute stability of global pattern formation and parallel memory storage by competitive neural networks},
  author={Cohen, Michael A and Grossberg, Stephen},
  journal={IEEE transactions on systems, man, and cybernetics},
  number={5},
  pages={815--826},
  year={1983},
  publisher={IEEE}
}

@book{lyons2007differential,
  title={Differential equations driven by rough paths: Ecole d'Et{\'e} de Probabilit{\'e}s de Saint-Flour XXXIV-2004},
  author={Lyons, Terry J and Caruana, Michael and L{\'e}vy, Thierry},
  year={2007},
  publisher={Springer}
}

@article{White1986,
  author = {White, J. G. and Southgate, E. and Thomson, J. N. and Brenner, S.},
  title = {The structure of the nervous system of the nematode Caenorhabditis elegans},
  journal = {Philosophical Transactions of the Royal Society of London. Series B, Biological Sciences},
  volume = {314},
  number = {1165},
  pages = {1--340},
  year = {1986},
  doi = {10.1098/rstb.1986.0056},
  url = {https://royalsocietypublishing.org}
}

@inproceedings{dao2022hungry,
  title={Hungry Hungry Hippos: Towards Language Modeling with State Space Models},
  author={Dao, Tri and Fu, Daniel Y and Saab, Khaled K and Thomas, Armin W and Rudra, Atri and R{\'e}, Christopher},
  booktitle={International Conference on Learning Representations (ICLR)},
  year={2023}
}

@inproceedings{wang2022pretraining,
  title={Pretraining Without Attention},
  author={Wang, Junxiong and Ma, Jing and Gu, Albert},
  booktitle={International Conference on Learning Representations (ICLR)},
  year={2023}
}

@article{gu2023mamba,
  title={Mamba: Linear-Time Sequence Modeling with Selective State Spaces},
  author={Gu, Albert and Dao, Tri},
  journal={arXiv preprint arXiv:2312.00752},
  year={2023}
}

@article{lieber2024jamba,
  title={Jamba: A Hybrid Transformer-Mamba Language Model},
  author={Lieber, Opher and others},
  journal={arXiv preprint arXiv:2403.19887},
  year={2024}
}

@article{wang2024mambabyte,
  title={MambaByte: Token-free Selective State Space Model},
  author={Wang, Junxiong and others},
  journal={arXiv preprint arXiv:2401.13660},
  year={2024}
}

@article{Cook2019,
  author    = {Cook, Steven J. and Jarrell, Travis A. and Brittin, Christopher A. and Wang, Yiming and Bloniarz, Adam E. and Yakovlev, Maksim A. and Nguyen,KC Q. and Austin, Leo T. J. and Kormish, David H. and Herndon, Ryan and Hall, David H. and Emmons, Scott W.},
  title     = {Whole-animal connectomes of both Caenorhabditis elegans sexes},
  journal   = {Nature},
  volume    = {571},
  number    = {7763},
  pages     = {63--71},
  year      = {2019},
  doi       = {10.1038/s41586-019-1352-7},
  url       = {https://doi.org}
}

@inproceedings{nguyen2022s4nd,
  title={S4ND: Modeling Images and Videos as Multidimensional Signals with State Spaces},
  author={Nguyen, Eric and Goel, Karan and Gu, Albert and Downs, Gordon and Shah, Preey and Dao, Tri and Baccus, Stephen and R{\'e}, Christopher},
  booktitle={Advances in Neural Information Processing Systems},
  volume={35},
  pages={2846--2861},
  year={2022}
}

@article{zhu2024vision,
  title={Vision Mamba: Efficient Visual Representation Learning with Bidirectional State Space Model},
  author={Zhu, Lianghui and Liao, Bencheng and Zhang, Qian and Wang, Xinlong and Liu, Wenyu and Wang, Xinggang},
  journal={arXiv preprint arXiv:2401.09417},
  year={2024}
}

@article{liu2024vmamba,
  title={VMamba: Visual State Space Model},
  author={Liu, Yue and Tian, Yunjie and Zhao, Yuzhong and Yu, Hongtian and Xie, Lingxi and Wang, Yaowei and Ye, Qixiang and Liu, Yunfan},
  journal={arXiv preprint arXiv:2401.10166},
  year={2024}
}

@inproceedings{zhu2025mambaml,
  title={Mambaml: Exploring state space models for multi-label image classification},
  author={Zhu, Xuelin and Liu, Jian and Cao, Jiuxin and Wang, Bing},
  booktitle={2025 IEEE/CVF International Conference on Computer Vision (ICCV)},
  pages={4743--4753},
  year={2025},
  organization={IEEE}

}

@inproceedings{johnson2012patient,
  title={Patient specific predictions in the intensive care unit using a Bayesian ensemble},
  author={Johnson, Alistair EW and Dunkley, Nic and Mayaud, Louis and Tsanas, Athanasios and Kramer, Andrew A and Clifford, Gari D},
  booktitle={Computing in Cardiology (CinC), 2012},
  pages={249--252},
  year={2012},
  organization={IEEE}
}

@inproceedings{goel2022sashimi,
  title={It's Raw! Audio Generation with State-Space Models},
  author={Goel, Karan and Gu, Albert and Donahue, Chris and R{\'e}, Christopher},
  booktitle={International Conference on Machine Learning (ICML)},
  pages={7616--7633},
  year={2022}
}

@inproceedings{debrouwer2019gru,
  title={{GRU-ODE-Bayes}: Continuous modeling of sporadically-observed time series},
  author={De Brouwer, Edward and Simm, Jaak and Arany, Adam and Moreau, Yves},
  booktitle={Advances in Neural Information Processing Systems (NeurIPS)},
  volume={32},
  pages={7379--7390},
  year={2019}
}

@inproceedings{rusch2021cornn,
  title={Coupled Oscillatory Recurrent Neural Network (coRNN): An accurate and (gradient) stable architecture for learning long time dependencies},
  author={Rusch, T. Konstantin and Mishra, Siddhartha},
  booktitle={International Conference on Learning Representations},
  year={2021},
  url={https://openreview.net/forum?id=F3s69XzWOia}
}

@inproceedings{dupont2019augmented,
  title={Augmented Neural ODEs},
  author={Dupont, Emilien and Doucet, Arnaud and Teh, Yee Whye},
  booktitle={Advances in Neural Information Processing Systems},
  volume={32},
  year={2019}
}

@article{bodnar2020second,
  title={On Second Order Behaviour in Augmented Neural ODEs},
  author={Norcliffe, Alexander and Bodnar, Cristian and Day, Ben and Simidjievski, Nikola and Li{\'o}, Pietro},
  booktitle={Advances in Neural Information Processing Systems},
  volume={33},
  pages={5911--5921},
  year={2020}
}

@inproceedings{jia2019neural,
  title={Neural jump stochastic differential equations},
  author={Jia, Junteng and Benson, Austin R},
  booktitle={Advances in Neural Information Processing Systems},
  volume={32},
  year={2019}
}

@inproceedings{de2024griffin,
  title={Griffin: Mixing Gated Linear Recurrences with Local Attention for Efficient Language Models},
  author={De, Soham and Smith, Samuel L. and Fernando, Anushan and Botev, Aleksandar and Cristian-Muraru, George and Gu, Albert and Haroun, Ruba and Berrada, Leonard and Chen, Yutian and Srinivasan, Srivatsan and Desjardins, Guillaume and Doucet, Arnaud and Budden, David and Teh, Yee Whye and Pascanu, Razvan and De Freitas, Nando and Gulcehre, Caglar},
  journal={arXiv preprint arXiv:2402.19427},
  year={2024}
}

@article{sun2023retentive,
  title={Retentive Network: A Successor to Transformer for Large Language Models},
  author={Sun, Yutao and Dong, Li and Huang, Shaohan and Ma, Shuming and Xia, Yuqing and Xue, Jilong and Wang, Jianyong and Wei, Furu},
  journal={arXiv preprint arXiv:2307.08621},
  year={2023}
}

@article{peng2023rwkv,
  title={RWKV: Reinventing RNNs for the Transformer Era},
  author={Peng, Bo and Alcaide, Eric and Anthony, Quentin and Albalak, Alon and Arcadinho, Samuel and Bao, Huan and Biswas, Rigoban and Cao, Arun and Chaim, Nathaniel and Chada, Nigel and others},
  journal={arXiv preprint arXiv:2305.13048},
  year={2023}
}

@article{yang2023gated,
  title={Gated Linear Attention Transformers with Hardware-Efficient Training},
  author={Yang, Songlin and Wang, Bailin and Shen, Yikang and Panda, Rameswar and Kim, Yoon},
  journal={arXiv preprint arXiv:2312.06635},
  year={2023}
}

@article{ren2024samba,
  title={Samba: Simple Hybrid State Space Models for Efficient Unlimited Context Language Modeling},
  author={Ren, Lingxiao and Liu, Yelong and Lu, Yutong and Zhou, Jingbo and Wei, Wei and Kong, Xian},
  journal={arXiv preprint arXiv:2406.07522},
  year={2024}
}

@inproceedings{greydanus2019hamiltonian,
  title={Hamiltonian Neural Networks},
  author={Greydanus, Samuel and Dzamba, Misko and Yosinski, Jason},
  booktitle={Advances in Neural Information Processing Systems},
  year={2019}
}

@article{cranmer2020lagrangian,
  title={Lagrangian Neural Networks},
  author={Cranmer, Miles and Greydanus, Samuel and Hoyer, Stephan and Battaglia, Peter and Spergel, David and Ho, Shirley},
  journal={arXiv preprint arXiv:2003.04630},
  year={2020}
}

@article{poli2021graph,
  title={Graph Neural Ordinary Differential Equations},
  author={Poli, Michael and Massaroli, Stefano and Park, Junyoung and Deaco, Attilio and Ermon, Stefano},
  journal={arXiv preprint arXiv:1911.07532},
  year={2021}
}

@inproceedings{tzen2019neural,
  title={Neural Stochastic Differential Equations: Deep Latent Gaussian Models in the Diffusion Limit},
  author={Tzen, Belinda and Raginsky, Maxim},
  journal={arXiv preprint arXiv:1905.09883},
  year={2019}
}

@inproceedings{neil2016phased,
  title={Phased LSTM: Accelerating Recurrent Network Training for Long or Event-based Sequences},
  author={Neil, Daniel and Pfeiffer, Michael and Liu, Shih-Chii},
  booktitle={Advances in Neural Information Processing Systems},
  year={2016}
}

@misc{morrill2021neuralroughdifferentialequations,
      title={Neural Rough Differential Equations for Long Time Series}, 
      author={James Morrill and Cristopher Salvi and Patrick Kidger and James Foster and Terry Lyons},
      year={2021},
      eprint={2009.08295},
      archivePrefix={arXiv},
      primaryClass={cs.LG},
      url={https://arxiv.org/abs/2009.08295}, 
}

@inproceedings{he2016deep,
  title={Deep residual learning for image recognition},
  author={He, Kaiming and Zhang, Xiangyu and Ren, Shaoqing and Sun, Jian},
  booktitle={Proceedings of the IEEE conference on computer vision and pattern recognition},
  pages={770--778},
  year={2016}
}

@inproceedings{ijcai2019p103,
  title = {ANODE: Unconditionally Accurate Memory-Efficient Gradients for Neural ODEs},
  author = {Gholaminejad, Amir and Keutzer, Kurt and Biros, George},
  booktitle = {Proceedings of the Twenty-Eighth International Joint Conference on Artificial Intelligence, {IJCAI-19}},
  pages = {730--736},
  year = {2019},
  doi = {10.24963/ijcai.2019/103}
}

@article{rackauckas2020universal,
  title={Universal differential equations for scientific machine learning},
  author={Rackauckas, Christopher and Ma, Yingbo and Martensen, Julius and Warner, Collin and Zubov, Kirill and Supekar, Rohit and Skinner, Dominic and Ramadhan, Ali},
  journal={arXiv preprint arXiv:2001.04385},
  year={2020}
}

@inproceedings{bilos2021neuralflows,
  title={Neural Flows: Efficient Alternative to Neural ODEs},
  author={Bilo{\v{s}}, Marin and Sommer, Johanna and Rangapuram, Syama Sundar and Januschowski, Tim and G{\"u}nnemann, Stephan},
  booktitle={Advances in Neural Information Processing Systems},
  volume={34},
  pages={21757--21771},
  year={2021}
}

@article{pechlivanidou2022zero,
  author    = {Pechlivanidou, Georgia and Karampetakis, Nicholas},
  title     = {Zero-order hold discretization of general state space systems with input delay},
  journal   = {IMA Journal of Mathematical Control and Information},
  volume    = {39},
  number    = {2},
  pages     = {708--730},
  year      = {2022},
  publisher = {Oxford University Press},
  doi       = {10.1093/imamci/dnac005}
}

@book{Pontryagin1962Mathematical,
  author    = {Pontryagin, L. S. and Boltyanskii, V. G. and Gamkrelidze, R. V. and Mishchenko, E. F.},
  title     = {The Mathematical Theory of Optimal Processes},
  publisher = {Interscience Publishers},
  year      = {1962},
  address   = {New York and London},
  note      = {Translated from the Russian by K. N. Trirogoff}
}

@article{physionet,
  title={PhysioBank, PhysioToolkit, and PhysioNet: components of a new research resource for complex physiologic signals},
  author={Goldberger, Ary L and others},
  journal={Circulation},
  volume={101},
  number={23},
  pages={e215--e220},
  year={2000},
  publisher={Am Heart Assoc}
}

@inproceedings{
  zhang2024cmamba,
  title={CMamba: Channel Correlation Enhanced State Space Models for Multivariate Time Series Forecasting}, 
  author={Chenglin Zhang and et al.},
  year={2024},
  eprint={2406.05316},
  archivePrefix={arXiv},
  primaryClass={cs.LG},
  url={https://arxiv.org/abs/2406.05316},
}

@misc{cai2024mambatsimprovedselectivestate,
      title={MambaTS: Improved Selective State Space Models for Long-term Time Series Forecasting}, 
      author={Xiuding Cai and Yaoyao Zhu and Xueyao Wang and Yu Yao},
      year={2024},
      eprint={2405.16440},
      archivePrefix={arXiv},
      primaryClass={cs.LG},
      url={https://arxiv.org/abs/2405.16440},
}

@misc{soydan2024s7selectivesimplifiedstate,
      title={S7: Selective and Simplified State Space Layers for Sequence Modeling}, 
      author={Taylan Soydan and Nikola Zubić and Nico Messikommer and Siddhartha Mishra and Davide Scaramuzza},
      year={2024},
      eprint={2410.03464},
      archivePrefix={arXiv},
      primaryClass={cs.LG},
      url={https://arxiv.org/abs/2410.03464},
}

@article{mozer2017discrete,
  title={Discrete Event, Continuous Time RNNs},
  author={Mozer, Michael C and Kazakov, Denis and Lindsey, Robert V},
  journal={arXiv preprint arXiv:1710.04110},
  year={2017}
}

@incollection{ornsteinuhlenbeck2014overview,
  title={Ornstein--Uhlenbeck processes and extensions},
  author={Maller, Ross A and M{\"u}ller, Gernot and Szimayer, Alex},
  journal={Handbook of financial time series},
  pages={421--437},
  year={2009},
  publisher={Springer}
}

@misc{zong2025accuracymemoryefficiencygeneralization,
      title={Accuracy, Memory Efficiency and Generalization: A Comparative Study on Liquid Neural Networks and Recurrent Neural Networks}, 
      author={Shilong Zong and Alex Bierly and Almuatazbellah Boker and Hoda Eldardiry},
      year={2025},
      eprint={2510.07578},
      archivePrefix={arXiv},
      primaryClass={cs.LG},
      url={https://arxiv.org/abs/2510.07578},
}

@inproceedings{jammal2025comparative,
  title={A Comparative Analysis of Liquid Neural Networks and Incremental Learning Approaches for Stock Market Prediction},
  author={Jammal, Hussein and Srour, Farah and Salman, Ali and Achkar, Roger},
  booktitle={2025 International Conference on Control, Automation and Diagnosis (ICCAD)},
  year={2025}
}

@article{Wahidi2025ScaleIN,
  title={Scale Is Not All You Need: Revisiting the Biomimetic Roots of Deep Learning to Overcome Fundamental Limitations},
  author={Malek Wahidi and Anthony Rizk and Rodrigue Imad},
  journal={IEEE Access},
  year={2025},
  volume={13},
  pages={125537-125569},
  doi={10.1109/ACCESS.2025.3589514}
}

@inproceedings{bhethanabhotla2024Mamba4Cast,
  title     = {Mamba4Cast: Efficient Zero-Shot Time Series Forecasting with State Space Models},
  author    = {Bhethanabhotla, Sathya Kamesh and Swelam, Omar and Siems, Julien and Salinas, David and Hutter, Frank},
  booktitle = {NeurIPS 2024 TSALM Workshop},
  year      = {2024}
}

@inproceedings{das2024decoderonly,
  title={A decoder-only foundation model for time-series forecasting},
  author={Das, Abhimanyu and Kong, Weihao and Sen, Rajat and Zhou, Yichen},
  booktitle={Proceedings of the 41st International Conference on Machine Learning},
  pages={10148--10167},
  year={2024},
  series={PMLR},
  volume={235}
}

@article{rasul2024lagllama,
  title={Lag-Llama: Towards Foundation Models for Probabilistic Time Series Forecasting},
  author={Rasul, Kashif and others},
  journal={arXiv preprint arXiv:2310.08278},
  year={2024}
}

@book{szeg1939orthogonal,
  title={Orthogonal polynomials},
  author={Szeg, Gabor},
  volume={23},
  year={1939},
  publisher={American Mathematical Soc.}
}

@article{Bekolay2014,
  title={Nengo: a Python tool for building large-scale functional brain models},
  author={Bekolay, Trevor and Bergstra, James and Hunsberger, Eric and DeWolf, Travis and Stewart, Terrence C and Rasmussen, Daniel and Choo, Xuan and Voelker, Aaron Russell and Eliasmith, Chris},
  journal={Frontiers in neuroinformatics},
  volume={7},
  pages={48},
  year={2014},
  publisher={Frontiers Media SA}
}

@inproceedings{zhuang2020adaptive,
  title={Adaptive Checkpoint Adjoint Method for Gradient Estimation in Neural ODE},
  author={Zhuang, Juntang and Dvornek, Nicha and Li, Xiaoxiao and Tatikonda, Sekhar and Papademetris, Xenophon and Duncan, James},
  booktitle={Proceedings of the International Conference on Machine Learning (ICML)},
  volume={119},
  pages={11639--11649},
  year={2020}
}

@article{farsang2024liquid,
  title={Liquid Resistance Liquid Capacitance Networks},
  author={Farsang, M{\'o}nika and Neubauer, Sophie A. and Grosu, Radu},
  journal={arXiv preprint arXiv:2403.08791},
  year={2024}
}

@article{Zhu2024HyperConnections,
  title     = {Hyper-Connections},
  author    = {Defa Zhu and Hongzhi Huang and Zihao Huang and Yutao Zeng and Yunyao Mao and Banggu Wu and Qiyang Min and Xun Zhou},
  journal   = {arXiv preprint arXiv:2409.19606},
  year      = {2024},
  url       = {https://arxiv.org/abs/2409.19606}
}

@article{nguyen2026muonssm,
  title={MuonSSM: Orthogonalizing State Space Models for Sequence Modeling},
  author={Nguyen, Thai-Khanh and Vo, Ngoc-Bich-Uyen and Vo, Thieu N and Nguyen, Tan M and Pham, Cuong},
  journal={arXiv preprint arXiv:2606.30461},
  year={2026}
}

@inproceedings{oh2024stable,
  title={Stable neural stochastic differential equations in analyzing irregular time series data},
  author={Oh, YongKyung and Lim, Dongyoung and Kim, Sungil},
  booktitle={International Conference on Learning Representations},
  volume={2024},
  pages={38231--38262},
  year={2024}
}

@article{mathieu2023geometric,
  title={Geometric neural diffusion processes},
  author={Mathieu, Emile and Dutordoir, Vincent and Hutchinson, Michael and De Bortoli, Valentin and Teh, Yee Whye and Turner, Richard},
  journal={Advances in Neural Information Processing Systems},
  volume={36},
  pages={53475--53507},
  year={2023}
}

@article{brandle2026continuous,
  title={Continuous-time piecewise-linear recurrent neural networks},
  author={Br{\"a}ndle, Alena and Eisenmann, Lukas and G{\"o}tz, Florian and Durstewitz, Daniel},
  journal={arXiv preprint arXiv:2602.15649},
  year={2026}
}

@article{mauel2026foundation,
  title={Foundation inference models for ordinary differential equations},
  author={Mauel, Maximilian and H{\"u}bers, Johannes R and Berghaus, David and Seifner, Patrick and Sanchez, Ramses J},
  journal={arXiv preprint arXiv:2602.08733},
  year={2026}
}

@article{shen2026enhancing,
  title={Enhancing Irregular Time Series Forecasting with Continuous-Time Modeling Framework},
  author={Shen, Tianen and Li, Zhengyu and Li, Yutong and Qiu, Xiangfei and Wu, Xingjian and Yang, Bin and Hu, Jilin},
  journal={arXiv preprint arXiv:2607.28035},
  year={2026}
}

@article{mizuguchi2026shippo,
  title={SHiPPO: Recurrent Memory with Transported Polynomial Projections},
  author={Mizuguchi, Tomoya and Kim, Bum Jun},
  journal={arXiv preprint arXiv:2607.03055},
  year={2026}
}

@article{jacot2018neural,
  title={Neural tangent kernel: Convergence and generalization in neural networks},
  author={Jacot, Arthur and Gabriel, Franck and Hongler, Cl{\'e}ment},
  journal={Advances in neural information processing systems},
  volume={31},
  year={2018}
}

@article{raissi2019physics,
  title={Physics-informed neural networks: A deep learning framework for solving forward and inverse problems involving nonlinear partial differential equations},
  author={Raissi, Maziar and Perdikaris, Paris and Karniadakis, George E},
  journal={Journal of Computational physics},
  volume={378},
  pages={686--707},
  year={2019},
  publisher={Elsevier}
}

@article{li2020fourier,
  title={Fourier neural operator for parametric partial differential equations},
  author={Li, Zongyi and Kovachki, Nikola and Azizzadenesheli, Kamyar and Liu, Burigede and Bhattacharya, Kaushik and Stuart, Andrew and Anandkumar, Anima},
  journal={arXiv preprint arXiv:2010.08895},
  year={2020}
}

@article{lechner2020neural,
  title={Neural circuit policies enabling auditable autonomy},
  author={Lechner, Mathias and Hasani, Ramin and Amini, Alexander and Henzinger, Thomas A and Rus, Daniela and Grosu, Radu},
  journal={Nature Machine Intelligence},
  volume={2},
  number={10},
  pages={642--652},
  year={2020},
  publisher={Nature Publishing Group UK London}
}

@article{razzaq2023neural,
  title={Neural Circuit Policies Imposing Visual Perceptual Autonomy: W. Razzaq, M. Hongwei},
  author={Razzaq, Waleed and Hongwei, Mo},
  journal={Neural Processing Letters},
  volume={55},
  number={7},
  pages={9101--9116},
  year={2023},
  publisher={Springer}
}

@article{razzaq2024neural,
  title={Neural Circuit Policies for Virtual Character Control: W. Razzaq, K. Raza},
  author={Razzaq, Waleed and Raza, Kashif},
  journal={Neural Processing Letters},
  volume={56},
  number={3},
  pages={188},
  year={2024},
  publisher={Springer}
}

@article{kim2024continuous,
  title={Continuous-time linear positional embedding for irregular time series forecasting},
  author={Kim, Byunghyun and Lee, Jae-Gil},
  journal={arXiv preprint arXiv:2409.20092},
  year={2024}
}

@inproceedings{zhang2025semi,
  title={Semi-implicit neural ordinary differential equations},
  author={Zhang, Hong and Liu, Ying and Maulik, Romit},
  booktitle={Proceedings of the AAAI Conference on Artificial Intelligence},
  volume={39},
  number={21},
  pages={22416--22424},
  year={2025}
}

@article{lee2021parameterized,
  title={Parameterized neural ordinary differential equations: Applications to computational physics problems},
  author={Lee, Kookjin and Parish, Eric J},
  journal={Proceedings of the Royal Society A},
  volume={477},
  number={2253},
  pages={20210162},
  year={2021},
  publisher={JSTOR}
}

@inproceedings{wang2024neural,
  title={Neural structure learning with stochastic differential equations},
  author={Wang, Benjie and Jennings, Joel and Gong, Wenbo},
  booktitle={International Conference on Learning Representations},
  volume={2024},
  pages={25306--25318},
  year={2024}
}

\end{document}